\documentclass{article}

\usepackage{iclr2027_conference,times}

\usepackage{amsmath,amsfonts,bm}

\def\eqref#1{equation~\ref{#1}}

\def\1{\bm{1}}

\DeclareMathAlphabet{\mathsfit}{\encodingdefault}{\sfdefault}{m}{sl}
\SetMathAlphabet{\mathsfit}{bold}{\encodingdefault}{\sfdefault}{bx}{n}

\usepackage{url}
\usepackage{graphicx}
\usepackage{booktabs}
\usepackage{tabularx}
\usepackage{float}
\usepackage{hyperref}
\newcounter{adagepaalgorithm}

\title{AdaGEPA: Adaptive Feedback Allocation for Reflective Prompt Optimization}
\author{%
\makebox[\dimexpr\textwidth-2\tabcolsep\relax][c]{%
\begin{tabular}[t]{c}
\textbf{Junyang Chen}$^{1}$ \qquad
\textbf{Zecheng Wang}$^{1}$ \qquad
\textbf{Jingbang Chen}$^{1,2,\dagger}$ \\[8pt]
{\mdseries $^{1}$The Chinese University of Hong Kong, Shenzhen} \\
{\mdseries $^{2}$Shenzhen Loop Area Institute} \\[10pt]
{\mdseries\small $^{\dagger}$Corresponding author.}
\end{tabular}%
}%
}

\iclrfinalcopy

\begin{document}

\maketitle
\fancyhead[L]{}

\begin{abstract}
Prompt optimization improves the performance of language-model systems on
downstream tasks by refining their prompts. Classical methods evaluate prompts
on task examples and use the resulting feedback to guide prompt revisions through
reflection. However, when feedback selection does not account for the
prompt's weaknesses, these revisions may improve performance on selected examples
without yielding broader task improvements. To address this issue, we propose
AdaGEPA, an adaptive feedback-allocation method that uses the prompt's
performance and task structure to select examples for the next prompt revision.
Our method replaces at most one example in each feedback minibatch to target an
identified weakness while preserving the remaining feedback context. Across
our main experiments on six
downstream benchmarks, AdaGEPA achieves higher mean validation scores than
non-adaptive feedback selection under matched rollout budgets. AdaGEPA also
finds high-performing prompts earlier across several tasks. In the initial
Schema-Guided Dialogue (SGD) study, its half-budget prompts outperform the
non-adaptive baseline's full-budget prompts in joint goal accuracy on new
dialogues from services seen and unseen during search. Overall, our findings
highlight the potential of adaptive
feedback allocation to improve both the effectiveness and rollout-budget efficiency
of reflective prompt optimization.
\end{abstract}

\section{Introduction}

Prompt optimization adapts language-model systems to downstream tasks by
refining their instructions without updating model weights
\citep{yang2024opro,pryzant2023automatic}. These tasks include task-oriented
dialogue \citep{rastogi2020sgd,budzianowski2018multiwoz}, mathematical
reasoning \citep{dekoninck2026matharena}, instruction following
\citep{pyatkin2025ifbench}, and multi-hop verification
\citep{jiang2020hover}. Beyond scalar outcomes, task executions can provide
richer evidence, including reasoning traces, tool outputs, and evaluator
feedback that can reveal why an output failed. Reflective prompt optimizers
use this evidence to guide subsequent prompt revisions.
\citet{agrawal2025gepa} introduced GEPA, a reflective prompt optimizer that
maintains a pool of candidate prompts and uses natural-language reflection
to revise a selected parent based on feedback from a minibatch of task
examples. Because the examples in that minibatch determine which evidence
shapes the next prompt revision, their selection can affect both the revised
candidate and the subsequent search trajectory. This raises a natural
question: how should feedback examples be selected to address the current
prompt's weaknesses and improve performance beyond the selected minibatch?

The GEPA default shuffles feedback examples without targeting the selected
parent's weaknesses. A natural
alternative is to resample the entire minibatch to favor examples with larger
loss gaps, repeated failures, or underrepresented task groups. However, our
initial experiments reveal a gap between early progress and the final search
outcome: although this strategy can improve early candidate quality and
produce more candidates that improve on their parents, it does not
consistently improve the best validation score found by the end of search.
One challenge is that reflection turns feedback from a small minibatch into
a prompt revision that affects behavior across the task. A revision that
fixes the selected errors may leave other errors unresolved or weaken
behavior that was already correct. This tension echoes two established
observations: example choice and ordering matter in curriculum learning,
while a shared update can help one behavior but harm another in multi-task
learning \citep{bengio2009curriculum,yu2020gradient}. These observations
suggest that difficulty and coverage alone do not reliably identify feedback
that leads to broader task improvements. We therefore study how task
structure and the parent prompt's performance can guide feedback selection
toward revisions that benefit examples beyond the minibatch.

\begin{figure}[t]
    \centering
    \includegraphics[width=\textwidth]{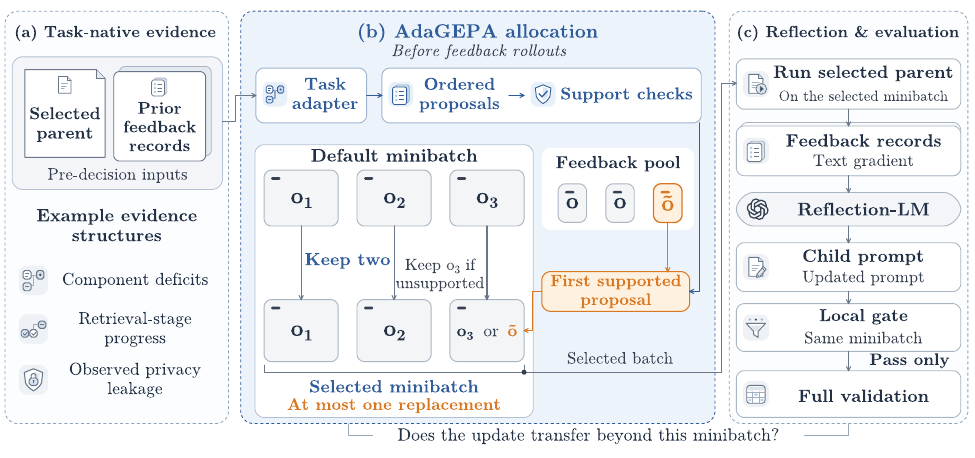}
    \caption{AdaGEPA adapts the feedback minibatch before reflection.
    (a) Cached evaluation records provide task-specific signals about the
    selected parent and search state. (b) A task adapter prioritizes eligible
    proposals and replaces at most one example in the default minibatch. If
    none passes the support checks, the default minibatch is retained.
    (c) The selected minibatch supplies feedback for the next prompt revision.
    The resulting candidate is compared with its parent on that minibatch and
    proceeds to full validation if its local score improves.}
    \label{fig:aft-gepa-overview}
\end{figure}

In this paper, we study pre-reflection feedback allocation and propose
AdaGEPA, a lightweight method that uses cached evaluation records and task
structure to select examples for the next prompt revision.
Figure~\ref{fig:aft-gepa-overview} shows how AdaGEPA makes this choice after
parent selection and before feedback generation. Depending on the task, this structure
may describe service categories, evaluation components, or intermediate
execution stages. Rather than resampling the entire minibatch, AdaGEPA
replaces at most one example in the default minibatch. Retaining the other
examples preserves most of the default feedback context while adding targeted
evidence about the prompt's weaknesses. If no eligible replacement is
available, AdaGEPA uses the default minibatch.

For example, in dialogue state tracking, cached evaluations can reveal
whether the current parent struggles with intent prediction or slot-value
tracking. A task-specific adapter can then propose an example whose task
profile is aligned with those weaknesses. The replacement is applied only
if it satisfies the task-specific eligibility and scheduling rules.
The selected examples then provide feedback for the prompt revision. The
resulting candidate is evaluated on the same minibatch and, if it improves
on the parent, proceeds to full validation to determine whether it improves
the best score found so far.

Our contributions are summarized as follows:
\begingroup
\setlength{\leftmargini}{1em}
\begin{itemize}
    \item We formalize pre-reflection feedback allocation as a distinct
    decision in reflective prompt search: after parent selection, the optimizer
    chooses which examples inform reflection. We define its immediate utility
    as the expected one-step gain in the best-so-far validation score.

    \item We propose AdaGEPA, a lightweight method that uses task structure
    and cached evaluations to target a prompt's weaknesses through at most one
    minibatch replacement, without adding language-model calls or altering
    subsequent reflection and evaluation.

    \item Across six downstream benchmarks, AdaGEPA yields higher mean final
    validation scores than default feedback selection at matched budgets and
    finds high-scoring prompts earlier. In the initial Schema-Guided Dialogue
    (SGD) study, prompts found using half the budget outperform full-budget
    baseline prompts on new dialogues from seen and unseen services.

    \item Through task-specific controls and trajectory analyses, we trace
    feedback choices through candidate evaluation, best-score improvements,
    and the parent-child path to the final winner (winner lineage). This shows how early candidate
    improvements relate to final outcomes and whether they persist as search
    continues.
\end{itemize}
\endgroup

\section{Related Work}

\paragraph{Textual feedback and prompt optimization.}
Prompt optimizers differ in how they use evaluation evidence. OPRO uses earlier
candidates and objective values to guide new proposals. EvoPrompt and
Promptbreeder combine LM-generated mutations with task-level fitness, while
MIPRO uses stochastic minibatches and a surrogate objective to optimize
instructions and demonstrations
\citep{yang2024opro,guo2024evoprompt,fernando2023promptbreeder,opsahlong2024mipro}.
Other methods use feedback to shape prompt revisions directly.
ProTeGi derives natural-language critiques from minibatch examples, TextGrad
propagates textual feedback through compound computation graphs, and GEPA
reflects on task-native feedback and execution trajectories within a
Pareto-based search
\citep{pryzant2023automatic,yuksekgonul2024textgrad,agrawal2025gepa}.
Our work focuses on the decision before reflection: which examples should
inform the next prompt revision. We then examine whether that revision
improves performance beyond the selected examples.

\paragraph{Data selection in prompt search.}
Data selection enters prompt search at several decision points. During
optimization, APEX uses prompt-lineage history and solvability to choose
mutation data, and rank sensitivity to choose candidate-evaluation data. Its
$2\times2$ ablation examines these choices separately \citep{wang2026apex}.
TwinBandit jointly selects challenging instances and mutation strategies
through two bandits \citep{park2025twinbandit}, whereas IPOMP updates the
candidate-evaluation subset as optimization proceeds
\citep{dong2025ipomp}. Other methods select data before the search begins.
SESS constructs a representative submodular evaluation subset under OPRO
\citep{nian2026sess}, and p1 filters user prompts by separating prompt-quality
variation from response stochasticity \citep{gao2026p1}. For reflective
updates, AdaGEPA uses the prompt's performance and task structure to select
the examples that guide the next prompt revision.

\paragraph{Downstream outcomes and search control.}
Downstream outcomes can guide individual edits or broader search decisions.
Learning to Self-Evolve rewards context edits by their improvement on a
holdout set and trains an editing policy with reinforcement learning
\citep{chen2026learning}. ReASearch lets an agent choose prompt and program
revisions, minibatches, evaluation budgets, and stopping decisions
\citep{li2026optimizer}. RLMOpt and OEO also place candidate revision,
evaluation, and search continuation under the optimizer's control
\citep{satheesha2026rlmopt,xue2026rethinking}. DIVE instead evolves reusable
skill artifacts through diversity-driven search \citep{xiong2026dive}. We
focus on an earlier, narrower decision: which examples inform reflection. We
examine whether the resulting candidate improves the best full-validation
score and whether it or one of its descendants becomes the final selected
prompt.

\section{Preliminaries}
\label{sec:preliminaries}

\subsection{Prompt Optimization under a Rollout Budget}
\label{sec:prompt-optimization-rollout-budget}

We study prompt optimization for a compound AI system under a task-level
rollout budget. We use $\Phi$ to denote a candidate system instantiated with a
particular prompt configuration. Model weights and all non-prompt components
remain fixed across candidates \citep{agrawal2025gepa}. A task instance is a
pair $(x,m)$, where $x$ is the system input and $m$ contains the information
used to evaluate its output. The task metric
$\mu(\Phi(x),m)\in[0,1]$ returns a scalar score, and the feedback function
$\mu_f$ supplements it with task-native textual feedback for reflection.

We measure this budget in rollouts. A rollout consists of running a candidate
$\Phi$ on one task instance and evaluating its output with $\mu$. The
optimization budget $B$ limits how many of these task-level evaluations may be
performed. Calls to models and tools, including the language model used for
reflection, are counted separately.

\subsection{Reflective Search with Scheduled Feedback}
\label{sec:gepa-search-default-allocation}

During development search, the optimizer uses a feedback set
$\mathcal D_{\mathrm{feedback}}$ and a validation set
$\mathcal D_{\mathrm{pareto}}$. At the start of each epoch, the examples in
$\mathcal D_{\mathrm{feedback}}$ are randomly shuffled and then used
sequentially in minibatches. We call each scheduled use of a task example a
\emph{feedback occurrence}, so repeated uses of the same example count as
distinct occurrences. Running a candidate on the selected occurrences produces
execution records and evaluator feedback for reflection.
$\mathcal D_{\mathrm{pareto}}$ supports full candidate evaluation, parent
selection, and winner selection.

At proposal step $t$, $\mathcal P_t$ contains the candidates that have
completed full evaluation on $\mathcal D_{\mathrm{pareto}}$. Parent selection
uses their instance-level scores, so a candidate that performs best on some
instances can still be selected as a parent even if it does not have the
highest average score
\citep{agrawal2025gepa}.

After an existing candidate $\Phi_k$ is selected as the parent, the default
feedback schedule supplies the next $b$ occurrences to form $\mathcal M_t^0$.
Running the parent on these occurrences produces execution records and
evaluator feedback. The Reflection-LM uses this information to revise one
prompt component of the parent candidate, producing a new candidate
$\Phi'$, called the child. The
resulting child is then evaluated on the same minibatch as its parent. If its
average score is higher than the parent's, it is fully evaluated on
$\mathcal D_{\mathrm{pareto}}$ and added to the candidate pool. Parent choice
and prompt revision therefore adapt during search, but the feedback minibatch
does not depend on the selected parent's current state.
Appendix~\ref{app:gepa-default-transition} provides additional details of this
search transition.

\subsection{Pre-Reflection Feedback Allocation}
\label{sec:pre-reflection-feedback-allocation}

Feedback allocation occurs after parent selection but before any rollout is
executed on the next feedback minibatch. The chosen occurrences therefore
determine which execution records and evaluator feedback serve as input to the
Reflection-LM. We call this decision \emph{pre-reflection feedback allocation}.
At proposal step $t$, let $\xi_t$ denote the pre-reflection search state,
including the selected parent, the default minibatch $\mathcal M_t^0$, cached
evaluation records, and the state of the feedback schedule. Let
$\mathbb A(\xi_t)$ denote the feedback-allocation actions available in this
state. An action $a_t\in\mathbb A(\xi_t)$ selects $b$ occurrences and produces
$\mathcal M_t(a_t)$. The default action $a_0$ leaves $\mathcal M_t^0$
unchanged, so $\mathcal M_t(a_0)=\mathcal M_t^0$. Across actions, the selected
parent, minibatch size, rollout cap, and all subsequent search rules remain
fixed. The optimizer chooses the action using only information available in
$\xi_t$, prior to generating the child or observing any later search outcome.

\section{Methods}
\label{sec:method}

\subsection{Motivation and Design Rationale}
\label{sec:allocation-design-rationale}
\label{sec:diagnosing-feedback-allocation}
\label{sec:local-repair-frontier-progress}
\label{sec:difficulty-transfer}

Feedback selection should consider not only whether the resulting prompt
improves the examples used for reflection, but also how it performs on the
full validation set. The local gate admits a child when it improves on its
parent over the selected minibatch \citep{agrawal2025gepa}. However, a child
that passes this gate need not outperform its parent under full validation
(Appendix~\ref{app:proposal-alignment-audit}). Even an improvement over the
parent may leave the best-so-far score unchanged. Therefore, feedback
allocation must distinguish local repair, parent-level improvement, and
frontier progress.

One simple alternative is to resample the entire minibatch using loss gaps,
repeated failures, and coarse task coverage. Our design studies show that this
history-and-coverage rule can improve early candidate quality, but its gains do
not consistently persist as search continues
(Appendices~\ref{app:sgd-full-minibatch-history-coverage}
and~\ref{app:livebench-full-minibatch-precursor}).
Therefore, difficulty and coverage can identify promising feedback, but cannot
tell us whether the resulting revision will advance the best-so-far validation
score.

To address this limitation, our proposed method, AdaGEPA, aims to combine
task structure with cached evaluations of the selected parent. Depending
on the task, this structure may describe behaviors, components, or stages,
while the cached records reveal where the parent performs poorly. A task
adapter uses this evidence to propose one replacement for the default
minibatch. By keeping the remaining examples unchanged, this one-slot design
preserves most of the default feedback context while directing one position
toward a task-relevant weakness.

\subsection{Immediate Frontier Utility}
\label{sec:frontier-transfer-utility}

We define the immediate utility of a feedback choice by how much the resulting
prompt revision advances the best validation score found so far. Let
$\bar S(\Phi)$ denote the average score of candidate $\Phi$ on the full validation set
$\mathcal D_{\mathrm{pareto}}$, and let $H_t$ be the highest score in the
candidate pool at the start of proposal step $t$:
\[
\begin{aligned}
    \bar S(\Phi)
    &=
    \frac{1}{|\mathcal D_{\mathrm{pareto}}|}
    \sum_{(x_i,m_i)\in\mathcal D_{\mathrm{pareto}}}
    \mu\!\left(\Phi(x_i),m_i\right),
    &
    H_t
    &=
    \max_{\Phi\in\mathcal P_t}\bar S(\Phi).
\end{aligned}
\]
Given a feedback choice $a\in\mathbb A(\xi_t)$, let $H_{t+1}^{(a)}$ be the
best-so-far score after applying $a$ in state $\xi_t$ and completing the
resulting proposal transition. We then define its one-step frontier gain,
expected gain, and advantage over the default action $a_0$ as
\[
\begin{aligned}
    g_t(a)
    &=
    H_{t+1}^{(a)}-H_t,
    &
    U(a\mid\xi_t)
    &=
    \mathbb E\!\left[g_t(a)\mid\xi_t\right],
    &
    \operatorname{Adv}(a\mid\xi_t)
    &=
    U(a\mid\xi_t)-U(a_0\mid\xi_t).
\end{aligned}
\]
Because reflection and candidate evaluation are stochastic, the same feedback
choice can yield different frontier gains even from the same state.
Accordingly, $U(a\mid\xi_t)$ denotes the expected frontier gain over this
randomness, while $\operatorname{Adv}(a\mid\xi_t)$ compares that expectation
with the expected gain under the default action $a_0$.

\subsection{Bounded Feedback Allocation}
\label{sec:task-native-bounded-allocation}
\label{sec:legal-actions-search-invariants}

AdaGEPA limits each action to at most one replacement, preserving most of the
default feedback context. Let the default minibatch $\mathcal M_t^0$ have size
$b$. The first $b-1$ occurrences remain fixed, and only the final occurrence
may be replaced:
\[
\begin{aligned}
    \mathcal M_t(a_0)
    &= (o_{t,1}^0,\ldots,o_{t,b-1}^0,o_{t,b}^0),
    &
    \mathcal M_t(a)
    &= (o_{t,1}^0,\ldots,o_{t,b-1}^0,\tilde o_{t,b}(a)).
\end{aligned}
\]
The replacement is chosen in three steps.
\textbf{(1) Propose.} A task adapter uses the evidence available in $\xi_t$ to
nominate possible replacement occurrences and rank them according to a fixed
priority rule.
\textbf{(2) Check.} A replacement proposal is supported only if the proposed occurrence
belongs to $\mathcal D_{\mathrm{feedback}}$, is distinct from every occurrence
in the default minibatch, and satisfies the adapter-specific schedule
constraints. Proposals targeting the same occurrence are treated as a single
action.
\textbf{(3) Select.} AdaGEPA selects the first supported replacement proposal in priority
order. If none is supported, it applies the default action $a_0$ and leaves
$\mathcal M_t^0$ unchanged.
Feedback selection does not consume the task-level rollout budget. Before a
proposal step begins, the runner verifies that enough budget remains to compare the
parent and child on the minibatch and to fully evaluate the child if it passes
the local gate.

Algorithm~\ref{alg:adagepa-selection} summarizes this shared selection
procedure. The task adapter $A$ supplies an ordered list of replacement proposals, while
the support check and default fallback are shared across tasks.

\par\smallskip
\noindent\begin{minipage}{\linewidth}
\small
\refstepcounter{adagepaalgorithm}\label{alg:adagepa-selection}
\begin{tabularx}{\linewidth}{@{}r>{\raggedright\arraybackslash}X@{}}
\toprule
\multicolumn{2}{@{}l}{\textbf{Algorithm \theadagepaalgorithm: AdaGEPA feedback selection}} \\
\midrule
\multicolumn{2}{@{}p{\linewidth}@{}}{\hangindent=4.8em\hangafter=1%
\makebox[4.8em][l]{\makebox[3.7em][c]{\textbf{Input}}\textbf{:}}selected parent $\Phi_k$, default minibatch $\mathcal M_t^0$,
cached records $C$, feedback set $\mathcal D_{\mathrm{feedback}}$,
schedule state $q$, frozen task adapter $A$.} \\
\multicolumn{2}{@{}p{\linewidth}@{}}{\hangindent=4.8em\hangafter=1%
\makebox[4.8em][l]{\makebox[3.7em][c]{\textbf{Output}}\textbf{:}}selected batch $\mathcal M$ and updated schedule state $q'$.} \\
\addlinespace[2pt]
1 & $L \leftarrow A.\mathrm{propose}(\Phi_k,\mathcal M_t^0,C,\mathcal D_{\mathrm{feedback}},q)$ \\
2 & \textbf{for each} distinct occurrence $o$ in $L$, in adapter priority order: \\
3 & \quad\textbf{if} $\mathrm{Supported}(o,\mathcal M_t^0,\mathcal D_{\mathrm{feedback}},q)$: \\
4 & \qquad $(\mathcal M,q')\leftarrow\mathrm{ReplaceFinal}(\mathcal M_t^0,o,q)$ \\
5 & \qquad\textbf{return} $(\mathcal M,q')$ \\
6 & \textbf{return} $(\mathcal M_t^0,q)$ \\
\bottomrule
\end{tabularx}
\end{minipage}
\par\smallskip

Feedback selection uses only information already available in $\xi_t$ and adds no model
calls. The parent then produces reflection records on the selected minibatch.
The child is evaluated on this minibatch and proceeds to full validation only
after passing the local gate.

\subsection{Task-Specific Proposal Rules}
\label{sec:schema-informed-proposal-ordering}

Task adapters convert cached evaluation records, task metadata, and allocation
history into prioritized replacement proposals. Depending on the task, the
proposal rule can be a deterministic rule, a replay policy over eligible
feedback-bearing occurrences, or a model fitted before evaluation. In each
case, the evidence schema, proposal semantics, priority rule, and any fitted
parameters are fixed before evaluation. At each search step, the adapter
applies this fixed configuration to the selected parent's cached evaluations
and the occurrences currently available under the feedback schedule.

For example, if cached evaluation records show that the selected parent
performs poorly on a particular behavior, a task adapter can propose an
eligible occurrence associated with that behavior. When several occurrences
qualify, a fixed task-specific rule ranks them using the available search
evidence. The relevant task structure can represent services, evaluation
components, or stages. Appendix~\ref{app:task-protocols} summarizes the
task-specific evidence schemas and proposal rules used in our experiments.

\section{Experiments}
\label{sec:experiments}

\subsection{Experimental Setup}
\label{sec:experimental-setup}
\label{sec:evaluation-design}

\textbf{Evaluation design.}
Development-search curves track when each method discovers high-quality
candidates and how the performance gap evolves with the rollout budget. We
compare AdaGEPA with the GEPA default at matched checkpoints by evaluating the
candidate saved by each method at the same rollout budget. In the initial SGD
study, we also compare AdaGEPA at $B=560$
with the full-budget GEPA default. For SGD, MultiWOZ, and HoVer, we further
evaluate frozen candidates on new examples (search-external panels), without
further prompt optimization.

\textbf{Tasks and common configuration.}
Our experiments separate design studies from evaluations of the final method.
Full-minibatch history-and-coverage studies on SGD, AIME, and
LiveBench-Math~\citep{white2025livebench} test whether these signals are
sufficient to guide feedback selection. We then evaluate AdaGEPA's bounded
one-slot intervention across six benchmarks: dialogue state tracking (SGD and
MultiWOZ), multi-hop verification (HoVer), instruction following (IFBench),
privacy-sensitive generation (PUPA), and algorithmic code generation (TACO).
The main paired experiments use feedback minibatches of three occurrences.
We use \texttt{gpt-4.1-mini} (Task-LM) to execute task examples and
\texttt{gpt-5.1} (Reflection-LM) to generate prompt revisions. Within each
comparison, the two arms
share the initial candidate, rollout cap, and search procedure. They differ only in feedback
selection. Matched ablations and controls examine the contributions of one-slot
replacement, task-specific targeting, proposal-pool construction, and ordering. Task
configurations, evaluation protocols, and additional results appear in
Appendices~\ref{app:implementation-protocol}
and~\ref{app:additional-results}.

\textbf{Metrics and statistics.}\label{sec:metrics-statistics-integrity}
Each task is evaluated using its task-specific metric. In the main AdaGEPA
results, endpoint scores are reported as percentages and paired differences as
percentage points. For dialogue state tracking, Balanced DST combines state
and dialogue-component correctness, while joint goal accuracy (JGA) requires
the complete dialogue state to be correct. HoVer reports exact
supporting-document retrieval, IFBench reports the rate of responses that pass
all instruction verifiers, PUPA averages response quality and privacy
preservation, and TACO reports mean problem pass rate. For development
searches, Gain-AUC measures the area between the best-so-far curve and its
initial score over the normalized rollout budget. We pair runs by optimization
seed and report mean scores and mean paired differences in the main tables.
Appendix~\ref{app:statistics-integrity} defines the metrics,
repeated-evaluation aggregation, and run-inclusion rules.
Appendix~\ref{app:additional-results} reports median paired differences,
win/tie/loss counts, uncertainty intervals, and per-run results.

\subsection{Main Results}
\label{sec:results-sealed-transfer}

\newcommand{\resultsheadercenter}[1]{\raisebox{6.8pt}[0pt][0pt]{#1}}
\newcommand{\resultspaircenter}[1]{\raisebox{-4.4pt}[0pt][0pt]{#1}}
\newcommand{\resultsfourrowcenter}[1]{\raisebox{-13.2pt}[0pt][0pt]{#1}}
\newcommand{\resultspanelcenter}[1]{\raisebox{4pt}[0pt][0pt]{#1}}
\newcommand{\resultsblockcenter}[1]{\raisebox{-9.6pt}[0pt][0pt]{#1}}
\newcommand{\resultsblockncenter}[1]{\raisebox{-5pt}[0pt][0pt]{#1}}

\begin{figure}[t]
    \centering
    \includegraphics[width=0.98\textwidth]{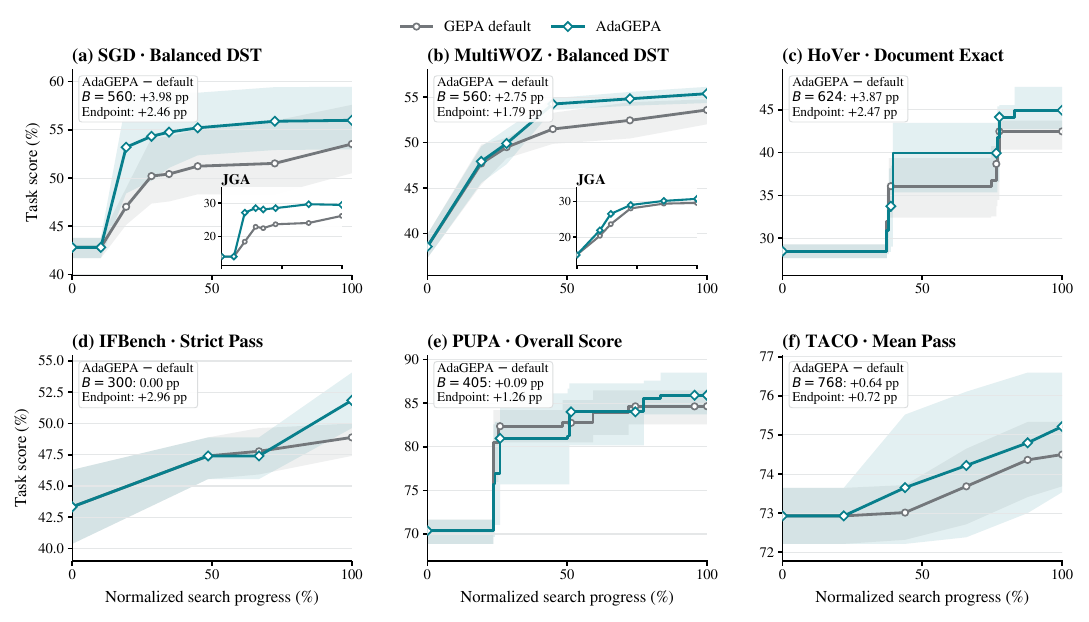}
    \caption{Development-search trajectories across six benchmarks. Curves show
    mean best-so-far task scores over normalized search progress, from the
    shared initial candidate at 0\% to the final checkpoint at 100\% within
    each task; shading shows 95\% seed-resampling intervals. Each panel uses
    the metric named in its title. The SGD and MultiWOZ insets show JGA, which
    requires the complete dialogue state to be correct. Callouts report mean
    paired differences (AdaGEPA minus GEPA default) at the labeled earlier
    checkpoint and endpoint while retaining the corresponding absolute $B$
    values. The SGD panel shows the initial study. HoVer and PUPA use step
    curves because candidates become available at seed-dependent budgets.}
    \label{fig:development-search-trajectories}
    \vspace{-3pt}
\end{figure}

The main experiments address two questions: whether adaptive feedback
selection helps search discover higher-scoring prompts under the same rollout
budget, and whether the resulting prompts retain their gains on examples not
used during search. Figure~\ref{fig:development-search-trajectories} traces
how best-so-far candidate quality changes over search, while
Table~\ref{tab:development-main-results} summarizes the earlier-checkpoint and
endpoint comparisons. In development search, AdaGEPA reached a higher mean
endpoint than the GEPA default across all six benchmarks. The clearest early
gains
appeared on SGD, MultiWOZ, and HoVer, while the smaller improvements on
IFBench, PUPA, and TACO emerged later. Two additional SGD studies also showed a clear early mean advantage, although the endpoint gaps gradually narrowed as the rollout budget increased further.

\begin{table}[t]
    \centering
    \scriptsize
    \setlength{\tabcolsep}{3.2pt}
    \renewcommand{\arraystretch}{1.10}
    \caption{Development-search candidate quality. Entries are mean task
    scores over optimization seeds, reported as percentages. Parentheses give
    paired mean differences (AdaGEPA minus GEPA default) in percentage points.
    Earlier checkpoints and endpoints are task-specific and listed by their
    absolute $B$ values. The final column reports the paired difference in
    budget-normalized Gain-AUC over each seed's development-search trajectory.}
    \label{tab:development-main-results}
    \begin{tabular}{@{}cccccc|ccc|c@{}}
        \toprule
        & & & \multicolumn{3}{c}{\textbf{Earlier checkpoint}}
        & \multicolumn{3}{c}{\textbf{Endpoint}} & \\
        \cmidrule(lr){4-6}\cmidrule(lr){7-9}
        \multicolumn{1}{c}{\resultsheadercenter{\textbf{Task / study}}} &
        \multicolumn{1}{c}{\resultsheadercenter{\textbf{Metric}}} &
        \resultsheadercenter{$\boldsymbol{n}$} &
        $\boldsymbol{B}$ & \multicolumn{1}{c}{\textbf{GEPA default}} &
        \multicolumn{1}{c}{\textbf{AdaGEPA ($\boldsymbol{\Delta}$)}} &
        $\boldsymbol{B}$ &
        \multicolumn{1}{c}{\textbf{GEPA default}} &
        \multicolumn{1}{c}{\textbf{AdaGEPA ($\boldsymbol{\Delta}$)}} &
        \resultsheadercenter{$\boldsymbol{\Delta}$ \textbf{Gain-AUC}} \\
        \midrule
        \resultspaircenter{SGD initial}
            & Balanced DST & \resultspaircenter{5} & \resultspaircenter{560} & 51.23 & \textbf{55.21} $(+3.98)$
            & \resultspaircenter{1120} & 53.53 & \textbf{55.99} $(+2.46)$ & \resultspaircenter{$+3.49$} \\
            & JGA & & & 23.65 & \textbf{28.46} $(+4.81)$
            & & 26.15 & \textbf{29.42} $(+3.27)$ & \\
        \addlinespace[1.5pt]
        \resultspaircenter{SGD extension}
            & Balanced DST & \resultspaircenter{5} & \resultspaircenter{560} & 51.80 & \textbf{53.59} $(+1.79)$
            & \resultspaircenter{1120} & \textbf{55.59} & 55.18 $(-0.40)$ & \resultspaircenter{$+0.73$} \\
            & JGA & & & 23.65 & \textbf{26.54} $(+2.88)$
            & & 27.50 & \textbf{28.27} $(+0.77)$ & \\
        \addlinespace[1.5pt]
        \resultspaircenter{SGD replication}
            & Balanced DST & \resultspaircenter{10} & \resultspaircenter{560} & 52.17 & \textbf{53.74} $(+1.58)$
            & \resultspaircenter{1120} & \textbf{55.35} & 55.33 $(-0.03)$ & \resultspaircenter{$+0.99$} \\
            & JGA & & & 24.42 & \textbf{26.54} $(+2.12)$
            & & \textbf{28.17} & \textbf{28.17} $(\phantom{+}0.00)$ & \\
        \addlinespace[1.5pt]
        \resultspaircenter{MultiWOZ}
            & Balanced DST & \resultspaircenter{5} & \resultspaircenter{560} & 51.50 & \textbf{54.25} $(+2.75)$
            & \resultspaircenter{1120} & 53.59 & \textbf{55.39} $(+1.79)$ & \resultspaircenter{$+1.48$} \\
            & JGA & & & 28.08 & \textbf{29.04} $(+0.96)$
            & & 29.62 & \textbf{30.77} $(+1.15)$ & \\
        \midrule
        HoVer
            & Document Exact & 5 & 624 & 36.07 & \textbf{39.93} $(+3.87)$
            & 1120 & 42.47 & \textbf{44.93} $(+2.47)$ & $+1.90$ \\
        IFBench
            & Strict pass & 5 & 300 & \textbf{47.41} & \textbf{47.41} $(\phantom{+}0.00)$
            & 560 & 48.89 & \textbf{51.85} $(+2.96)$ & $+1.23$ \\
        PUPA
            & Privacy--quality & 5 & 405 & 83.90 & \textbf{83.99} $(+0.09)$
            & 607 & 84.64 & \textbf{85.90} $(+1.26)$ & $-0.11$ \\
        TACO
            & Mean pass & 5 & 768 & 73.01 & \textbf{73.65} $(+0.64)$
            & 1424 & 74.50 & \textbf{75.21} $(+0.72)$ & $+0.45$ \\
        \bottomrule
    \end{tabular}
\end{table}

To test the second question, we evaluated frozen candidates from SGD,
MultiWOZ, and HoVer on examples not used during development search. The
clearest transfer result came from the initial SGD study. Using half the
optimization rollout budget, AdaGEPA found prompts that exceeded the GEPA
default's full-budget prompts by $7.69$ JGA points on new dialogues from
previously seen services (in-domain) and $5.19$ points on unseen services.
The two additional SGD studies also favored AdaGEPA at
$B=560$ on both panels, although the endpoint gaps narrowed as the GEPA
default continued to improve. Table~\ref{tab:key-results-contrasts} reports
both Balanced DST and the stricter JGA for SGD and MultiWOZ. On MultiWOZ,
AdaGEPA led by $1.56$ Balanced DST points at $B=560$ and $3.12$ points at
$B=1120$. HoVer also showed positive mean search-external differences of
$1.50$ Document Exact points at $B=624$ and $1.90$ points at $B=1120$.

\begin{table}[t]
    \centering
    \scriptsize
    \setlength{\tabcolsep}{2.9pt}
    \renewcommand{\arraystretch}{1.10}
    \caption{Search-external candidate quality. Entries are mean task scores
    over optimization seeds, reported as percentages. Parentheses give paired
    mean differences (AdaGEPA minus GEPA default) in percentage points.
    Earlier checkpoints and endpoints are listed by their absolute $B$ values.
    Dialogue-state-tracking rows report Balanced DST followed by JGA.}
    \label{tab:key-results-contrasts}
    \begin{tabular}{@{}ccccccc|ccc@{}}
        \toprule
        & & & & \multicolumn{3}{c}{\textbf{Earlier checkpoint}}
        & \multicolumn{3}{c}{\textbf{Endpoint}} \\
        \cmidrule(lr){5-7}\cmidrule(l){8-10}
        \multicolumn{1}{c}{\resultsheadercenter{\textbf{Task / study}}} &
        \multicolumn{1}{c}{\resultsheadercenter{\textbf{Panel}}} &
        \multicolumn{1}{c}{\resultsheadercenter{\textbf{Metric}}} &
        \resultsheadercenter{$\boldsymbol{n}$} &
        $\boldsymbol{B}$ &
        \multicolumn{1}{c}{\textbf{GEPA default}} &
        \multicolumn{1}{c}{\textbf{AdaGEPA ($\boldsymbol{\Delta}$)}} &
        $\boldsymbol{B}$ &
        \multicolumn{1}{c}{\textbf{GEPA default}} &
        \multicolumn{1}{c}{\textbf{AdaGEPA ($\boldsymbol{\Delta}$)}} \\
        \midrule
        \resultsfourrowcenter{SGD initial}
            & \resultspaircenter{In-domain} & Balanced DST
            & \resultsfourrowcenter{5} & \resultsfourrowcenter{560}
            & 50.04 & \textbf{56.95} $(+6.91)$
            & \resultsfourrowcenter{1120} & 51.47 & \textbf{56.85} $(+5.38)$ \\
            & & JGA & & & 22.31 & \textbf{31.15} $(+8.85)$
            & & 23.46 & \textbf{31.73} $(+8.27)$ \\
            & \resultspaircenter{Unseen service} & Balanced DST
            & & & 56.29 & \textbf{61.03} $(+4.74)$
            & & 57.33 & \textbf{59.69} $(+2.36)$ \\
            & & JGA & & & 28.08 & \textbf{34.04} $(+5.96)$
            & & 28.85 & \textbf{32.31} $(+3.46)$ \\
        \addlinespace[1.5pt]
        \resultsfourrowcenter{SGD extension}
            & \resultspaircenter{In-domain} & Balanced DST
            & \resultsfourrowcenter{5} & \resultsfourrowcenter{560}
            & 51.24 & \textbf{53.85} $(+2.61)$
            & \resultsfourrowcenter{1120} & 55.27 & \textbf{55.40} $(+0.13)$ \\
            & & JGA & & & 23.85 & \textbf{27.88} $(+4.04)$
            & & 28.08 & \textbf{29.62} $(+1.54)$ \\
            & \resultspaircenter{Unseen service} & Balanced DST
            & & & 56.21 & \textbf{59.88} $(+3.67)$
            & & 59.90 & \textbf{60.40} $(+0.50)$ \\
            & & JGA & & & 26.99 & \textbf{32.69} $(+5.71)$
            & & 32.12 & \textbf{32.88} $(+0.77)$ \\
        \addlinespace[1.5pt]
        \resultsfourrowcenter{SGD replication}
            & \resultspaircenter{In-domain} & Balanced DST
            & \resultsfourrowcenter{10} & \resultsfourrowcenter{560}
            & 51.05 & \textbf{54.24} $(+3.19)$
            & \resultsfourrowcenter{1120} & 54.20 & \textbf{55.24} $(+1.04)$ \\
            & & JGA & & & 23.08 & \textbf{28.65} $(+5.58)$
            & & 27.40 & \textbf{29.62} $(+2.21)$ \\
            & \resultspaircenter{Unseen service} & Balanced DST
            & & & 56.56 & \textbf{57.67} $(+1.11)$
            & & \textbf{59.04} & 58.79 $(-0.24)$ \\
            & & JGA & & & 27.88 & \textbf{29.90} $(+2.02)$
            & & \textbf{31.25} & 30.87 $(-0.38)$ \\
        \midrule
        \resultspaircenter{MultiWOZ}
            & \resultspaircenter{Search-external} & Balanced DST
            & \resultspaircenter{5} & \resultspaircenter{560}
            & 50.92 & \textbf{52.47} $(+1.56)$
            & \resultspaircenter{1120} & 51.21 & \textbf{54.33} $(+3.12)$ \\
            & & JGA & & & \textbf{26.92}
            & \textbf{26.92} $(\phantom{+}0.00)$
            & & 25.77 & \textbf{28.65} $(+2.88)$ \\
        \midrule
        HoVer &
            Search-external & Document Exact & 5
            & 624 & 40.32 & \textbf{41.82} $(+1.50)$
            & 1120 & 44.74 & \textbf{46.64} $(+1.90)$ \\
        \bottomrule
    \end{tabular}
    \par\smallskip
    \begin{minipage}{\textwidth}
    \footnotesize
    $n$ counts paired optimization runs. Panel construction and metric
    definitions appear in Appendix~\ref{app:statistics-integrity}; rollout
    checkpoints and full paired results appear in
    Appendices~\ref{app:rollout-budget-details} and
    \ref{app:search-transfer-results}.
    \end{minipage}
\end{table}

\subsection{Ablation Studies}
\label{sec:results-budget-retention}
\label{sec:results-conditional-portability}

We focus the main component ablations on SGD, which provides the most complete
set of matched controls. Controls on the remaining tasks examine the same
design choices under different feedback structures, with full results reported
in Appendix~\ref{app:allocation-controls}.

\begin{figure}[t]
    \centering
    \includegraphics[width=0.98\textwidth]
    {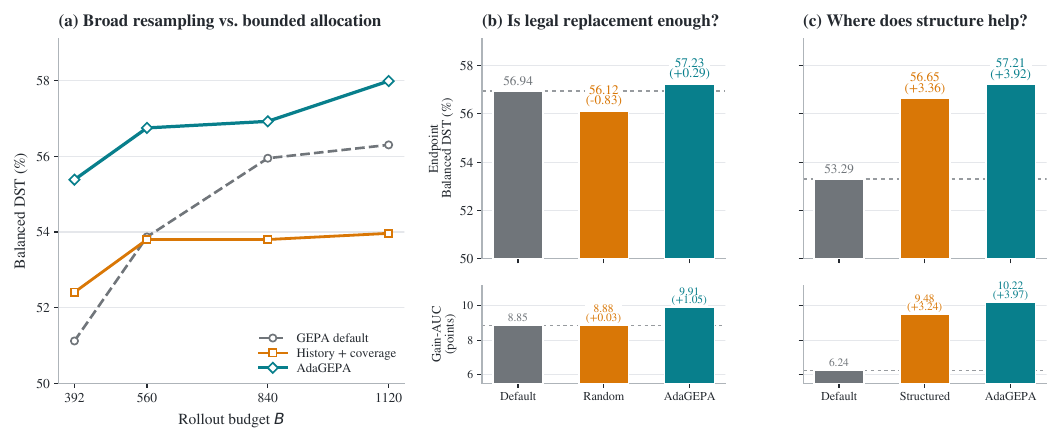}
    \caption{\textbf{Component ablations on SGD.}
    (a) Best-so-far development Balanced DST compares the GEPA default,
    full-minibatch history-and-coverage selection, and AdaGEPA using an
    ordering fitted from ten prior searches.
    (b) Endpoint Balanced DST at $B=1120$ (upper) and Gain-AUC (lower)
    compare the GEPA default, Random Legal, and AdaGEPA over ten paired
    optimization runs. Random Legal uniformly samples one legal replacement.
    (c) The same measures compare AdaGEPA with Structured Unordered, which
    uses the same supported replacement pool and fallback but omits fitted
    ordering. Parentheses show mean paired differences from the GEPA default.
    Full statistics and additional controls appear in
    Appendix~\ref{app:sgd-development-controls}.}
    \label{fig:sgd-core-ablation}
\end{figure}

We first examine how much of the feedback minibatch should be changed. On SGD,
full-minibatch history-and-coverage sampling improved early candidate quality
but stalled as the GEPA default continued to improve, motivating a more
bounded intervention (Figure~\ref{fig:sgd-core-ablation}a). On AIME, replacing
the full minibatch with a coarse selection rule left an
endpoint deficit of $1.6$ correct answers relative to the GEPA default.
Applying the same rule to only one slot reduced this gap to $0.4$. The one-slot
constraint therefore bounds the intervention by introducing targeted feedback
while preserving most of the default feedback context.

We next examine which components make a one-slot replacement useful. Random
Legal retains the one-slot constraint but samples uniformly from the legal
alternatives, testing whether any legal replacement is sufficient without
task-specific selection. Relative to the GEPA default, its mean Gain-AUC
differed by only $+0.03$ points, while its endpoint was $0.83$ points lower.
AdaGEPA exceeded Random Legal by $1.02$ Gain-AUC points and $1.12$ endpoint
points. These results suggest that bounding the intervention to one
replacement does not by itself improve search. Structured Unordered retains
AdaGEPA's supported proposal pool and fallback but omits fitted ordering. It
recovered most of AdaGEPA's mean improvement over the default, suggesting that
structuring the proposal space accounts for most of the observed gain. The
fitted ordering provided a smaller additional improvement that varied across
configurations. Full paired results and additional controls are reported in
Appendix~\ref{app:sgd-development-controls}.

Finally, we examine how task-specific structure should guide feedback
selection. On SGD, AdaGEPA refines service-level selection using intent and
slot deficits. Its mean unseen-service JGA exceeded the service-level-only
control by $4.81$ points at $B=560$ and $0.38$ points at $B=1120$
(Table~\ref{tab:sgd-service-only-control}). On MultiWOZ, service-aware routing
yielded a mean endpoint Balanced DST $2.15$ points above uniform selection of
unused examples, while fitted occurrence ordering within the selected service
provided no further gain. Across both tasks, task-specific proposal structure
was more effective than generic coverage alone, while finer fitted ordering
was not consistently beneficial.

\section{Discussion}
\label{sec:discussion}

The development curves suggest that feedback selection may matter most when
useful feedback is hard to reach within the search budget. AdaGEPA can find
useful candidates early, while its lead sometimes narrows as search continues.
One possible explanation is that AdaGEPA's task-specific feedback selection
brings relevant examples into reflection before the shuffled schedule reaches
them. This leads to a
testable hypothesis: holding the task and adapter fixed, AdaGEPA should provide the greatest benefit when the default schedule can cover only a small part of the feedback pool within the search budget.

Selecting a relevant example does not ensure that the resulting prompt revision
will remain useful through later search. On MultiWOZ, service-aware routing
produced higher development scores than uniform selection of unused examples,
while fitted within-service ordering did not preserve that gain. PUPA illustrates the
next hurdle: even when feedback from observed privacy failures was reused, most
resulting candidates did not appear in winner lineages. A child can enter the
candidate pool without setting a new best score and may later be
selected as the parent for another revision. Therefore, a fuller assessment of
feedback allocation should consider both the immediate frontier change and
whether the resulting child lies on the winner lineage
(Appendix~\ref{app:mechanism-funnel-counts}).

Across tasks, AdaGEPA operates at the same pre-reflection decision point and
uses task-specific evaluator signals to identify weaknesses of the parent
prompt and guide replacement proposals. Development search tracks when useful
candidates are discovered, and search-external evaluation tests the resulting
prompts on new examples. Our results suggest that adaptive feedback allocation
can accelerate candidate discovery when relevant feedback is scarce. Its
longer-term value depends on whether the resulting revisions generalize beyond
the selected examples and remain useful as search proceeds.

\section{Conclusion}

Reflective prompt optimization depends not only on how a prompt is revised but
also on which feedback examples shape that revision. We proposed AdaGEPA to
address this allocation problem using task structure and observed prompt
performance. Our analyses show why this choice matters: improvements on the
selected examples do not reliably advance the best score on the full
validation set. Across six benchmarks, AdaGEPA achieved higher mean validation
scores under matched rollout budgets and often found higher-scoring candidates
earlier in search. Search-external evaluations on SGD, MultiWOZ, and HoVer
further showed positive mean gains on new examples. These findings highlight
feedback allocation as a consequential design choice for reflective prompt
optimization, affecting both when useful prompts are found and whether their
gains extend to new examples. A promising next step is to account for
longer-term search value by selecting feedback that produces useful future
parents, even when the immediate candidate does not advance the frontier.

\subsection*{AI use statement}
We used generative AI tools to assist with writing and editing experimental
and analysis scripts, organizing and analyzing experiment records, identifying
potentially relevant literature, and restructuring, language editing, and
polishing of the manuscript. The authors reviewed all AI-assisted outputs,
checked reported numerical results against the corresponding experiment
records, and verified citations against their original sources. The authors
take responsibility for the final text, claims, code, and other artifacts.


\subsection*{Reproducibility statement}
Sections~\ref{sec:method} and~\ref{sec:experimental-setup} describe
AdaGEPA's feedback-selection procedure and evaluation design.
Appendix~\ref{app:proposal-alignment-audit} documents the local-to-full
proposal-alignment audit.
Appendix~\ref{app:implementation-protocol} specifies the task adapters, data
roles and evaluation panels, rollout budgets, model request settings, metrics,
and statistical procedures. Appendix~\ref{app:additional-results} provides
additional checkpoint and paired-seed results, controls, and resource
accounting. The accompanying code artifact contains an executable
reference implementation of the shared bounded one-slot feedback-allocation
contract and its GEPA-compatible batch sampler.

\bibliography{references}
\bibliographystyle{iclr2027_conference}

\clearpage
\appendix
\section{Local-to-Full Proposal Alignment}
\label{app:proposal-alignment-audit}

We examine local-to-full alignment across 458 proposal transitions from 60
development runs on AIME, LiveBench-Math, and IFBench. Each transition passed
the strict local gate and completed full evaluation within the common
$B\leq560$ rollout prefix. The runs include the GEPA default and several
historical adaptive policies, so this analysis characterizes proposal-level
alignment rather than the effect of a particular method.

For each proposal, the local change is the child-minus-parent score change on
the selected minibatch, while the full-set change is the corresponding change
on $\mathcal D_{\mathrm{pareto}}$. Because every minibatch contains three
examples, using score sums rather than averages does not change the sign or
ordering of the local changes. Correlations use raw score-sum changes for both
quantities. Because the
validation-set sizes differ across tasks, the pooled correlations are
descriptive. The Pearson and Spearman correlations were $0.076$ and $0.108$,
respectively, indicating weak local-to-full alignment.

\begin{table}[H]
    \centering
    \small
    \caption{Full-validation child-minus-parent outcomes among proposals that
    passed the local gate. Dataset rows partition the complete proposal set,
    while the GEPA default row is a cross-task subset.}
    \label{tab:proposal-alignment-audit}
    \begin{tabular}{cccccc}
        \toprule
        \textbf{Scope} & \textbf{Proposals} & \textbf{Positive} &
        \textbf{Zero} & \textbf{Negative} & \textbf{Positive rate} \\
        \midrule
        AIME & 142 & 77 & 16 & 49 & 54.23\% \\
        LiveBench-Math & 226 & 111 & 28 & 87 & 49.12\% \\
        IFBench & 90 & 45 & 10 & 35 & 50.00\% \\
        All proposals & 458 & 233 & 54 & 171 & 50.87\% \\
        \midrule
        GEPA default subset & 150 & 85 & 14 & 51 & 56.67\% \\
        \bottomrule
    \end{tabular}
\end{table}

Although every proposal passed the local gate, 233 of 458 improved on its
parent under full validation. The remaining 225 were unchanged or worse.
Together with the low correlations, these outcomes distinguish improvement on
the selected minibatch from progress on the full validation set. This analysis
follows realized search trajectories rather than forked alternatives and
therefore does not estimate the counterfactual advantage of different feedback
choices from the same state.

\section{Implementation and Evaluation Protocol}
\label{app:implementation-protocol}

\noindent\textbf{Roadmap.}
We first describe the shared search setting, default search transition,
allocation contract, and task-specific adapters. We then specify the
comparison controls and rollout budgets, followed by the evaluation,
statistical, model-request, and integrity protocols.

\subsection{Search Setting and Data Roles}
\label{app:runner-data-models}

All paired comparisons use the same GEPA search implementation, with candidate
merging disabled. AdaGEPA intervenes after parent selection and changes only
the feedback minibatch used for reflection. Within each comparison, the
methods share all remaining search and evaluation settings, including the
rollout budget and model request configuration. Search-external evaluations
use frozen checkpoint candidates without further prompt optimization.

During optimization, $\mathcal D_{\mathrm{feedback}}$ supplies examples and
evaluator feedback for reflection and local parent--child comparison, whereas
$\mathcal D_{\mathrm{pareto}}$ supports full candidate evaluation, parent
selection, and final winner selection.

Table~\ref{tab:optimization-repeats} summarizes the optimization runs and
saved-state sets used in the reported comparisons. Paired search arms start
from the same prompt and, in the main AdaGEPA comparisons, reuse the same
initial validation records.\footnote{The historical LiveBench-Math
full-minibatch study evaluates the initial state separately for each arm.}
Search-external evaluations reuse candidates from the corresponding searches
and therefore do not add optimization runs.

\begin{table*}[t]
\centering
\small
\setlength{\tabcolsep}{4pt}
\renewcommand{\arraystretch}{1.06}
\caption{Optimization runs and saved-state sets used in the reported
comparisons. For SGD, 5-formation and 10-formation identify the number of formation
searches used to fit the ordering. Same-state comparisons reuse states from
their source searches.}
\label{tab:optimization-repeats}
\begin{tabularx}{\textwidth}{
    @{}
    >{\centering\arraybackslash}p{0.16\textwidth}
    >{\centering\arraybackslash}X
    >{\centering\arraybackslash}p{0.31\textwidth}
    @{}
}
\toprule
\textbf{Task} & \textbf{Experiment} & \textbf{Repeats} \\
\midrule
SGD & Ordering formation sets & 5 / 10 searches \\
SGD & Initial study (5 formation searches) & 5 paired runs \\
SGD & Extension study (5 formation searches) & 5 paired runs \\
SGD & Replication study (10 formation searches) & 10 paired runs \\
SGD & History-and-coverage and formation controls & 5 paired runs \\
SGD & Random Legal three-arm control & 10 paired runs \\
SGD & Strong-prompt continuation & \mbox{5 paired continuations} \\
SGD & Same-state comparisons & \mbox{2 panels, each from 5 source runs} \\
MultiWOZ & Service-aware search and allocation controls & 5 paired runs \\
HoVer & Stage-progress replay & 5 paired runs \\
AIME & Full-minibatch and bounded controls & 5 paired runs \\
\mbox{LiveBench-Math} & Full-minibatch design study & 10 paired runs \\
IFBench & Verifier-allocation comparison & 5 paired runs \\
PUPA & Privacy-feedback reuse & 5 paired runs \\
TACO & Coarse-skill allocation & 5 paired runs \\
\bottomrule
\end{tabularx}
\end{table*}

\subsection{GEPA Search Transition}
\label{app:gepa-default-transition}

Before proposal step \(t\), the candidate pool \(\mathcal P_t\) contains all
candidates that have undergone full evaluation on
\(\mathcal D_{\mathrm{pareto}}\). GEPA samples a parent using its Pareto-based
selector \citep{agrawal2025gepa}. The selector prunes redundant candidates from
the parent-selection support and then samples each remaining candidate in proportion
to the number of validation instances on which it is tied for the best score.
This pruning affects only parent selection and does not remove candidates from
\(\mathcal P_t\).

The default sampler maintains an epoch-shuffled schedule over
\(\mathcal D_{\mathrm{feedback}}\). After parent selection, the next \(b\)
scheduled instances form the feedback minibatch \(\mathcal M_t^0\). Running the
parent \(\Phi_k\) on this minibatch produces outputs, scores, execution
trajectories, and task-native feedback. The Reflection-LM uses these records to
revise one prompt component of the parent and produce a child candidate
\(\Phi'\).

The parent and child are evaluated on the same occurrence-level minibatch. The
strict local gate advances the child to full validation only if its average
minibatch score exceeds that of the parent. The implementation compares the
corresponding score sums, which yields the same decision because both
candidates are evaluated on the same \(b\) occurrences. Under the configuration
used in this paper, a child that passes the local gate is fully evaluated on
\(\mathcal D_{\mathrm{pareto}}\) and added to the candidate pool. At the end of
search, the candidate with the highest average score on
\(\mathcal D_{\mathrm{pareto}}\) is returned.

\subsection{Shared Allocation Contract}
\label{app:shared-allocation-contract}

The six task-specific AdaGEPA adapters share the one-slot contract defined in
Section~\ref{sec:method}. AdaGEPA starts from a three-occurrence default
minibatch, retains the first two occurrences as anchors, and allows the task
adapter to propose a replacement only for the third. The proposal is executed
only if it passes the task-specific support check and scheduling rules.
Otherwise, the default minibatch is retained.

\subsection{Task-Specific Adapters}
\label{app:task-protocols}

The shared interface leaves the evidence schema and proposal rule task
specific. Table~\ref{tab:task-specific-adapter-names} lists the adapter names
used in the results, while Table~\ref{tab:frozen-adapter-mappings} summarizes
how each task maps evaluation evidence to a feedback proposal. The proposal
rules and any fitted parameters are fixed before evaluation.

\begin{table}[t]
\centering
\small
\setlength{\tabcolsep}{4pt}
\renewcommand{\arraystretch}{1.08}
\caption{Task-specific AdaGEPA adapters. The main results use the shared
AdaGEPA name; mechanism tables use shortened forms of these adapter names.}
\label{tab:task-specific-adapter-names}
\begin{tabular}{@{}cc@{}}
\toprule
\textbf{Benchmark} & \textbf{Task-specific adapter} \\
\midrule
SGD & Hierarchical service--component adapter \\
MultiWOZ & Service-aware routing adapter \\
HoVer & Stage-progress replay adapter \\
IFBench & Mean-deficit adapter \\
PUPA & Observed-deficit replay adapter \\
TACO & Coarse-skill adapter \\
\bottomrule
\end{tabular}
\end{table}

AIME and LiveBench-Math are reported as design studies rather than as part of
the six-benchmark adapter comparison. Table~\ref{tab:frozen-adapter-mappings}
includes AIME because it also contains a bounded one-slot comparison;
LiveBench-Math examines only full-minibatch allocation and is reported
separately.

\begin{table*}[t]
\centering
\small
\setlength{\tabcolsep}{3.5pt}
\renewcommand{\arraystretch}{1.15}
\renewcommand{\tabularxcolumn}[1]{m{#1}}
\caption{Task-specific evidence schemas and proposal rules for the bounded
one-slot comparisons. Full-minibatch design controls are summarized separately
in Table~\ref{tab:control-designs}.}
\label{tab:frozen-adapter-mappings}
\begin{tabularx}{\textwidth}{
    @{}
    >{\centering\arraybackslash}m{0.11\textwidth}
    m{0.23\textwidth}
    X
    >{\raggedright\arraybackslash}m{0.24\textwidth}
    @{}
}
\toprule
\multicolumn{1}{c}{\textbf{Task}} &
\multicolumn{1}{c}{\textbf{Evidence used}} &
\multicolumn{1}{c}{\textbf{Proposal rule}} &
\multicolumn{1}{c}{\textbf{Question examined}} \\
\midrule
\textbf{SGD} &
Search progress, service exposure, static task profiles, and selected-parent
component deficits. &
Frozen models rank legal service-level proposals and, when supported, refine
the choice within the selected service. &
Component-aware proposal ordering and search-external candidate quality. \\
\addlinespace
\textbf{MultiWOZ} &
Selected-parent component deficits, service exposure, and frozen service
metadata. &
A frozen service-aware rule ranks legal service-level proposals using search
context, intervention exposure, and candidate-wide DST features, then selects
an eligible occurrence within the chosen service. &
Service-aware routing versus generic coverage and fitted within-service
occurrence ordering. \\
\addlinespace
\textbf{HoVer} &
Intermediate retrieval progress, unresolved final outcomes, and replay
history. &
A fixed replay rule prioritizes an eligible occurrence whose trajectory made
retrieval progress but did not resolve the task. &
Stage-progress-guided feedback reuse. \\
\addlinespace
\textbf{AIME} &
Failure history, repeated exposure, and coarse topic clusters. &
The bounded rule applies the history-and-coverage signal to at most one
supported replacement. &
History-and-coverage allocation under a bounded one-slot intervention. \\
\addlinespace
\textbf{IFBench} &
Selected-parent verifier deficits and prior exposure under overlapping
objectives. &
The fixed mean-deficit rule prioritizes supported examples from weak verifier
groups. &
Feedback allocation under overlapping verifier objectives. \\
\addlinespace
\textbf{PUPA} &
Update type, prior feedback observations, and visible privacy-leakage signals. &
On privacy-facing updates, a fixed rule reuses eligible feedback with
previously observed privacy leakage. Other update types retain the default
minibatch. &
Feedback reuse and downstream candidate outcomes under coupled objectives. \\
\addlinespace
\textbf{TACO} &
Validation-supported skill labels and selected-parent skill deficits. &
A fixed rule prioritizes a supported example from an underperforming skill
group. &
Whether coarse skill identity provides sufficient resolution for feedback
allocation. \\
\bottomrule
\end{tabularx}
\end{table*}

The SGD instantiation uses fitted proposal ordering.
Table~\ref{tab:sgd-reference-adapter-configuration} summarizes its formation
evidence, frozen models, and runtime selection rule. All fitted quantities
remain fixed during evaluation.

\begin{table*}[t]
\centering
\small
\setlength{\tabcolsep}{4pt}
\renewcommand{\arraystretch}{1.1}
\renewcommand{\tabularxcolumn}[1]{m{#1}}
\caption{Compact configuration of the fitted SGD proposal ordering.}
\label{tab:sgd-reference-adapter-configuration}
\begin{tabularx}{\textwidth}{
    @{}
    >{\centering\arraybackslash}m{0.20\textwidth}
    >{\raggedright\arraybackslash}X
    @{}
}
\toprule
\multicolumn{1}{c}{\textbf{Component}} &
\multicolumn{1}{c}{\textbf{Specification}} \\
\midrule
Formation evidence &
The reported configurations use five or ten designated development searches
under the default feedback rule to collect proposal states and one-step targets. \\
\addlinespace
Ordering target &
The target is the observed one-step Balanced-DST frontier gain after a complete
proposal transition. Rejected or non-advancing proposals receive zero. \\
\addlinespace
Frozen models &
Two ridge models rank legal service-level proposals and eligible within-service
refinements using search progress, service exposure, static task
profiles, and selected-parent deficits. \\
\addlinespace
Runtime selection &
The service-level model ranks legal service-level proposals, and the refinement
model ranks eligible occurrences within the selected service. A distinct
refinement is used when available; otherwise the service-level proposal is
used, with the default minibatch as the final fallback. \\
\bottomrule
\end{tabularx}
\end{table*}

\subsection{Comparison Controls}
\label{app:comparison-controls}

Table~\ref{tab:control-designs} summarizes the controls used to separate action
capacity, replacement support, ordering, and task-specific targeting. The
APEX-style and SESS-style controls adapt ideas from their source systems
\citep{wang2026apex,nian2026sess}. They preserve the surrounding search
transition but are not full reproductions of those systems.

\begin{table*}[t]
\centering
\small
\setlength{\tabcolsep}{3.5pt}
\renewcommand{\arraystretch}{1.12}
\caption{Control designs and the factors they are intended to isolate.}
\label{tab:control-designs}
\begin{tabularx}{\textwidth}{
    @{}
    >{\centering\arraybackslash}m{0.20\textwidth}
    >{\raggedright\arraybackslash}m{0.19\textwidth}
    m{0.23\textwidth}
    >{\raggedright\arraybackslash}X
    @{}
}
\toprule
\multicolumn{1}{c}{\textbf{Control}} &
\multicolumn{1}{c}{\textbf{Preserved structure}} &
\multicolumn{1}{c}{\textbf{Changed factor}} &
\multicolumn{1}{c}{\textbf{Purpose}} \\
\midrule
SGD Random Legal &
One-slot capacity, the legal replacement pool, and the search schedule &
Samples uniformly from legal replacements without task-aware ordering &
Isolate the value of legal replacement alone. \\
\addlinespace
SGD Structured Unordered &
AdaGEPA's supported proposal set and one-slot capacity &
Selects from the same supported proposal set without fitted occurrence ordering &
Separate structured replacement support from fitted ordering. \\
\addlinespace
SGD service-level-only &
One-slot capacity, service-level targeting, and the search transition &
Omits within-service component refinement &
Isolate component-level targeting beyond service selection. \\
\addlinespace
MultiWOZ coverage-only &
One-slot capacity and prior exposure information &
Selects unused examples without task-specific service routing &
Compare generic unused-example coverage with service-aware routing. \\
\addlinespace
MultiWOZ fitted ordering &
One-slot capacity and service-aware proposal routing &
Ranks supported occurrences within the selected service &
Measure the value of fitted ordering beyond rank-neutral service-aware routing. \\
\addlinespace
AIME bounded one-slot &
The history-and-coverage signal and surrounding search transition &
Applies the same signal to at most one replacement instead of resampling the
full minibatch &
Separate selection direction from intervention capacity. \\
\addlinespace
IFBench all-pass-risk &
One-slot capacity, verifier feedback, and support checks &
Aggregates feedback by all-verifier failure risk rather than mean deficit &
Compare alternative aggregations of overlapping verifier feedback. \\
\addlinespace
PUPA coverage-only &
One-slot capacity and the surrounding search transition &
Selects unseen examples without observed leakage feedback &
Compare generic coverage with observed-feedback reuse. \\
\addlinespace
History and coverage (SGD, AIME, LiveBench-Math) &
The surrounding search transition and rollout accounting &
Resamples the full minibatch using difficulty, repetition, and coverage &
Test the earlier full-minibatch allocation hypothesis. \\
\addlinespace
SGD formation size &
The fitted adapter architecture and evaluation protocol &
Varies the number of development searches used to fit proposal ordering &
Measure sensitivity to the amount of formation evidence. \\
\addlinespace
APEX-style (SGD, AIME) &
The task-matched feedback capacity and search transition &
Prioritizes feedback using failure history &
Provide a neighboring adaptive mutation-data control. \\
\addlinespace
SESS-style (SGD) &
The search transition and epoch composition &
Uses a fixed representative schedule rather than parent-conditioned selection &
Compare adaptive allocation with static representativeness. \\
\bottomrule
\end{tabularx}
\end{table*}

The PUPA component-composition comparator changes candidate components rather
than feedback selection and is therefore reported separately.

\subsection{SGD Same-State Fork Protocol}
\label{app:sgd-same-state-protocol}

Each comparison restores a fixed search state $\xi_t$. Each distinct action is
evaluated using $K=3$ independent proposal transitions; when two arms select
the same action, they share those transitions. For a deterministic arm that
chooses action $a$, we estimate its immediate frontier gain and its contrast
with the default choice $a_0$ as
\[
\widehat U(a\mid\xi_t)
=\frac{1}{K}\sum_{r=1}^{K}\bigl(H_{t+1}^{(a,r)}-H_t\bigr),
\qquad
\widehat{\operatorname{Adv}}(a\mid\xi_t)
=\widehat U(a\mid\xi_t)-\widehat U(a_0\mid\xi_t).
\]
Random Legal independently draws an action from the legal replacement pool on
each repeat. Its estimate therefore averages over both the replacement
distribution and the randomness of the resulting proposal transitions.
Transition repeats are first averaged within each state, after which
state-level contrasts are aggregated within the source optimization run. The
resulting source-run means form the statistical units in the reported
summaries.

\begin{table}[t]
\centering
\small
\setlength{\tabcolsep}{4pt}
\renewcommand{\arraystretch}{1.08}
\renewcommand{\tabularxcolumn}[1]{m{#1}}
\caption{Same-state fork panels. Every selected state admits every arm
compared in its panel.}
\label{tab:sgd-same-state-protocol}
\begin{tabularx}{\columnwidth}{
    @{}
    m{0.20\columnwidth}
    >{\centering\arraybackslash}m{0.25\columnwidth}
    X
    m{0.12\columnwidth}
    @{}
}
\toprule
\multicolumn{1}{c}{\textbf{Panel}} &
\multicolumn{1}{c}{\textbf{Source states}} &
\multicolumn{1}{c}{\textbf{Actions compared}} &
\multicolumn{1}{c}{\textbf{Repeats}} \\
\midrule
Four-arm comparison &
15 states from five development runs &
GEPA default, Random Legal, Structured Unordered, and AdaGEPA (5 formation runs) &
3 per distinct action \\
\addlinespace
Ordering comparison &
15 states from a separate set of five development runs &
Structured Unordered and AdaGEPA (10 formation runs) &
3 per distinct action \\
\bottomrule
\end{tabularx}
\end{table}

The two panels use separate source states and are analyzed independently.
Results appear in Appendix~\ref{app:sgd-same-state-comparisons}.

\subsection{Rollout Budgets and Candidate Checkpoints}
\label{app:rollout-budget-details}

The rollout budget $B$ counts logical candidate--example evaluations at the
task level, rather than raw provider requests. A proposal step includes a local
parent--child comparison and may also trigger full validation when the child
passes the local gate. The runner begins a proposal step only when the
remaining budget can cover both stages if needed. The candidate saved at a
checkpoint is therefore selected only from candidates whose full validation is
complete by the stated budget.

Rollout caps and reported checkpoints follow the fixed protocol for each
benchmark. Because development-validation sizes and proposal schedules differ,
absolute $B$ values are benchmark-specific.
Table~\ref{tab:rollout-budget-protocols} lists the rollout cap,
development-validation size, and checkpoints used in the reported comparisons.

\begin{table}[t]
\centering
\small
\setlength{\tabcolsep}{4pt}
\renewcommand{\arraystretch}{1.1}
\caption{Rollout caps, development-validation sizes, and key checkpoints used
in the reported comparisons.}
\label{tab:rollout-budget-protocols}
\begin{tabular*}{\textwidth}{@{\extracolsep{\fill}}cc@{\hspace{5.5em}}cc@{}}
\toprule
\textbf{Task} &
\textbf{Rollout cap} $\boldsymbol{B}$ &
\mbox{\textbf{Development-validation size}} &
\textbf{Key reported checkpoints} \\
\midrule
SGD & 1120 & 104 & 104, 392, 560, 840, 1120 \\
MultiWOZ & 1120 & 104 & 104, 300, 392, 560, 840, 1120 \\
HoVer & 1120 & 300 & 300, 624, 1120 \\
AIME & 560 & 45 & 45, 150, 300, 336, 392, 560 \\
\mbox{LiveBench-Math} & 560 & 50 & 50, 150, 300, 560 \\
IFBench & 560 & 54 & 54, 300, 392, 560 \\
PUPA & 607 & 111 & 111, 405, 607 \\
TACO & 1424 & 256 & 256, 768, 1424 \\
\bottomrule
\end{tabular*}
\end{table}

Because $B$ is an upper bound, local rejection and budget reservation for
possible full validation can leave the realized number of task-level
evaluations below the cap.

\subsection{Evaluation Panels, Metrics, and Statistics}
\label{app:statistics-integrity}

Development results use each task's search validation data. Search-external
evaluations score frozen checkpoint candidates on separate examples that were
not used during search or candidate selection. SGD uses two 104-example panels:
a train-derived in-domain panel and an official-test panel covering 15 unseen
services. MultiWOZ uses a deterministic 104-dialogue panel from its official
test split. HoVer uses a fixed 253-example panel that excludes examples sharing
source IDs with the search data and reports the full 300-example panel as a
secondary evaluation. AIME divides its 90 development examples evenly between
feedback and validation. Table~\ref{tab:task-native-metrics} summarizes the
evaluation settings and metrics for all tasks.

\begin{table*}[t]
\centering
\small
\setlength{\tabcolsep}{4pt}
\renewcommand{\arraystretch}{1.08}
\caption{Evaluation panels and data roles. Search-external panels evaluate
frozen checkpoint candidates without further prompt optimization.}
\label{tab:task-native-metrics}
\begin{tabularx}{\textwidth}{
    @{}
    >{\centering\arraybackslash}m{0.16\textwidth}
    >{\centering\arraybackslash}p{0.28\textwidth}
    >{\centering\arraybackslash}X
    @{}
}
\toprule
\multicolumn{1}{c}{\textbf{Task}} &
\multicolumn{1}{c}{\textbf{Development evaluation}} &
\multicolumn{1}{c}{\textbf{Search-external evaluation}} \\
\midrule
SGD & Search validation & In-domain (104); unseen-service (104) \\
MultiWOZ & Search validation & Search-external panel (104; official-test split) \\
HoVer & Search validation & Search-external panel (253); full evaluation panel (300) secondary \\
AIME & Held-out development split & \multicolumn{1}{c}{---} \\
\mbox{LiveBench-Math} & Search validation & \multicolumn{1}{c}{---} \\
IFBench & Search validation & \multicolumn{1}{c}{---} \\
PUPA & Search validation & \multicolumn{1}{c}{---} \\
TACO & Search validation & \multicolumn{1}{c}{---} \\
\bottomrule
\end{tabularx}
\end{table*}

\begin{table*}[t]
\centering
\small
\setlength{\tabcolsep}{4pt}
\renewcommand{\arraystretch}{1.12}
\renewcommand{\tabularxcolumn}[1]{m{#1}}
\caption{Task metrics and their aggregation.}
\label{tab:task-metric-definitions}
\begin{tabularx}{\textwidth}{
    @{}
    >{\centering\arraybackslash}m{0.16\textwidth}
    >{\centering\arraybackslash}m{0.18\textwidth}
    X
    @{}
}
\toprule
\multicolumn{1}{c}{\textbf{Task}} &
\multicolumn{1}{c}{\textbf{Metric}} &
\multicolumn{1}{c}{\textbf{Definition and aggregation}} \\
\midrule
SGD / MultiWOZ &
Balanced DST; JGA &
Balanced DST averages a per-example score with weight $1/2$ on exact
joint-state correctness and $1/6$ each on active-intent correctness,
requested-slot F1, and slot-value F1. Checkpoint candidates are selected by
Balanced DST. JGA is the fraction of examples with a completely correct
dialogue state; both metrics are reported for the same frozen candidate. \\
\addlinespace
HoVer &
Document Exact &
Fraction of examples for which the cumulative retrieval after the final hop
covers all three normalized gold supporting-document titles. \\
\addlinespace
AIME &
Formatted-answer accuracy &
Fraction of responses containing the formatted reference answer
(`\texttt{\#\#\# <integer>}'); the number of solved problems is also reported. \\
\addlinespace
\mbox{LiveBench-Math} &
Task-specific score &
Mean score assigned by the pinned task-specific deterministic scorers; the
number of full-credit problems is also reported. \\
\addlinespace
IFBench &
Strict pass rate &
Mean of binary item scores, where an item receives one only when the response
passes every instruction-specific verifier. \\
\addlinespace
PUPA &
Privacy--quality score &
Let $F$ indicate that the generated response is judged at least as good as the
reference, and let $R$ indicate the reverse comparison. The per-example quality score is
$q=\mathbf{1}[F\lor(F=R)]$, with $\mathbf{1}[F\land\neg R]$ as a stricter
secondary measure. If $\ell$ is the fraction of unique private-information
entries leaked into the request sent to the external model, privacy is
$p=1-\ell$. The overall score is $(q+p)/2$, averaged across validation examples. \\
\addlinespace
TACO &
Mean problem pass rate; strict pass rate &
Mean problem pass rate averages the fraction of tests passed for each problem.
Strict pass rate counts a problem as correct only when all of its tests pass. \\
\bottomrule
\end{tabularx}
\end{table*}

For development protocols evaluated at multiple rollout checkpoints, let
$H(B)$ denote the quality of the best budget-complete candidate available by
rollout $B$. Over the evaluated budget range $[B_0,B_1]$, we summarize
progress above the initial checkpoint as
\[
    \operatorname{Gain\text{-}AUC}
    =
    \frac{1}{B_1-B_0}
    \int_{B_0}^{B_1}
    \!\left(H(B)-H(B_0)\right)\,dB.
\]
This budget-normalized area captures both how early improvements appear and
how large they are. For metrics reported as proportions, Gain-AUC is multiplied
by 100 and reported in points.\footnote{AIME and LiveBench-Math retain their
original correct-answer scales.}

The optimization run, indexed by its seed, is the statistical unit. We first
compute the difference between AdaGEPA and its reference method within each
paired seed, then summarize the differences by their mean, median,
win/tie/loss (W/T/L) counts, and a 95\% uncertainty interval. When a frozen
candidate has repeated evaluation draws, we average those scores before
computing the paired difference. For bootstrap intervals, five-run analyses
enumerate all paired-seed resamples, whereas ten-run analyses use 100,000
samples drawn with a fixed random seed.\footnote{The PUPA comparisons retain
the $t$ intervals specified in their frozen analyses.} Reported aggregates
include all completed, protocol-conforming runs at the stated checkpoint.

\subsection{Model Requests and Integrity}
\label{app:model-requests-integrity}

\begin{table*}[t]
\centering
\scriptsize
\setlength{\tabcolsep}{4pt}
\renewcommand{\arraystretch}{1.08}
\caption{Model request configurations for the reported experiments.
\(T\) denotes temperature, \(L\) the maximum output-token setting,
\(R\) the configured retry count, and \(\tau\) the request timeout in
seconds.}
\label{tab:model-request-configuration}
\begin{tabularx}{\textwidth}{
    @{}
    >{\centering\arraybackslash}m{0.14\textwidth}
    >{\raggedright\arraybackslash}X
    >{\raggedright\arraybackslash}X
    @{}
}
\toprule
\multicolumn{1}{c}{\textbf{Task}} &
\multicolumn{1}{c}{\textbf{Task-LM or task evaluator}} &
\multicolumn{1}{c}{\textbf{Reflection-LM}} \\
\midrule
SGD, MultiWOZ &
\texttt{openai/gpt-4.1-mini};
\(T=1\), \(L=1024\), \(R=5\), \(\tau=240\) &
\texttt{openai/gpt-5.1};
\(T=1\), \(L=16000\), \(R=3\), \(\tau=600\) \\
HoVer &
\texttt{openai/gpt-4.1-mini};
\(T=1\), \(L=512\), \(R=2\), \(\tau=600\) &
\texttt{openai/gpt-5.1};
\(T=1\), \(L=4096\), \(R=2\), \(\tau=600\) \\
AIME &
\texttt{openai/gpt-4.1-mini};
\(T=1\), \(L=16000\), \(R=5\), \(\tau=240\) &
\texttt{openai/gpt-5.1};
\(T=1\), \(L=16000\), \(R=3\), \(\tau=600\) \\
LiveBench-Math &
\texttt{openai/gpt-4.1-mini};
\(T=1\), \(L=16000\), \(R=3\), \(\tau=600\) &
\texttt{openai/gpt-5.1};
\(T=1\), \(L=16000\), \(R=3\), \(\tau=600\) \\
IFBench &
\texttt{openai/gpt-4.1-mini};
\(T=1\), \(L=16000\), \(R=5\), \(\tau=240\) &
\texttt{openai/gpt-5.1};
\(T=1\), \(L=16000\), \(R=3\), \(\tau=600\) \\
PUPA &
\texttt{openai/gpt-4.1-mini};
task \(T=1,L=1024\), evaluator \(T=0,L=128\);
\(R=5\), \(\tau=180\) &
\texttt{openai/gpt-5.1};
\(T=1\), \(L=4096\), \(R=5\), \(\tau=180\) \\
TACO &
\texttt{openai/gpt-4.1-mini};
\(T=1\), \(L=16000\), \(R=3\), \(\tau=600\) &
\texttt{openai/gpt-5.1};
\(T=1\), \(L=16000\), \(R=3\), \(\tau=600\) \\
\bottomrule
\end{tabularx}
\end{table*}

None of the reported configurations explicitly sets \texttt{top\_p}. Model
names are given in normalized provider/model form. The recorded requests
identify each model endpoint but do not include a provider-issued immutable
snapshot identifier.

Subsequent Task-LM and Reflection-LM executions are independent across arms,
allowing their search trajectories to diverge. Quality summaries include all
completed, protocol-conforming runs that reach the reported checkpoint and
respect the budget and data boundaries. Failed runs are not replaced.
Post-search analyses use cached records without modifying the recorded
trajectories or scores. Logical rollout budgets and physical resources are
accounted for separately. The rollout budget $B$ follows the task-level
protocol, while provider retries, Reflection-LM calls, tokens, configured
cost, and wall-clock time are recorded independently. Equal rollout caps
therefore need not imply equal physical resource use.

\section{Additional Empirical Evidence}
\label{app:additional-results}

\noindent\textbf{Roadmap.}
We first provide detailed search and search-external results for the
comparisons reported in the main text. We then examine what drives feedback
allocation and where its value persists or disappears. The appendix concludes
with a cross-task synthesis and resource accounting.

\subsection{Search and Search-External Results}
\label{app:search-transfer-results}

This section expands the search and search-external comparisons reported in
the main text. It presents the three tasks with search-external panels---SGD,
MultiWOZ, and HoVer---together with the SGD formation-prompt comparisons.
Development-only controls for the remaining tasks appear in
Appendices~\ref{app:allocation-controls} and
\ref{app:mechanism-funnel-counts}, followed by the task-level synthesis in
Appendix~\ref{app:task-outcome-summary}.

\subsubsection{Initial SGD Study}
\label{app:sgd-checkpoint-curve}

\begin{figure*}[t]
    \centering
    \includegraphics[width=\textwidth]{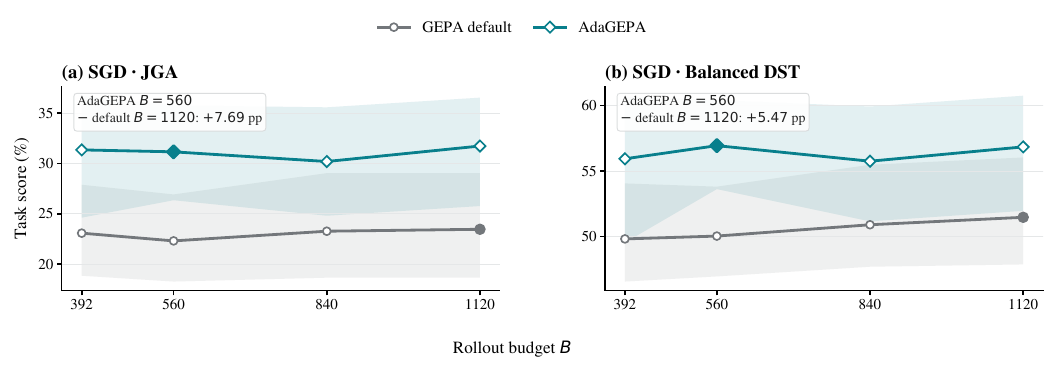}
    \caption{Search-external in-domain SGD quality across frozen rollout
    checkpoints. Curves show means over five paired optimization runs on the
    104-example panel, with 95\% seed-resampling intervals. Filled markers
    identify AdaGEPA at $B=560$ and the GEPA default at $B=1120$. Annotations
    report their paired mean differences.}
    \label{fig:sgd-sealed-budget-curve}
\end{figure*}

Figure~\ref{fig:sgd-sealed-budget-curve} shows in-domain candidate quality
across the frozen checkpoints, while
Table~\ref{tab:sgd-original-ood-checkpoint-curve} reports the corresponding
results on the unseen-service panel.

\begin{table*}[t]
\centering
\small
\setlength{\tabcolsep}{4.1pt}
\renewcommand{\arraystretch}{1.08}
\caption{Search-external SGD candidate quality across frozen checkpoints on
the 104-dialogue panel covering all 15 unseen services. Values are means over
five paired optimization runs.}
\label{tab:sgd-original-ood-checkpoint-curve}
\begin{tabular}{@{}lrrrrrrrr@{}}
\toprule
& \multicolumn{4}{c}{\textbf{JGA (\%)}} &
  \multicolumn{4}{c}{\textbf{Balanced DST (\%)}} \\
\cmidrule(lr){2-5}\cmidrule(lr){6-9}
\multicolumn{1}{c}{\textbf{Arm}} &
$\boldsymbol{B392}$ & $\boldsymbol{B560}$ & $\boldsymbol{B840}$ & $\boldsymbol{B1120}$ &
$\boldsymbol{B392}$ & $\boldsymbol{B560}$ & $\boldsymbol{B840}$ & $\boldsymbol{B1120}$ \\
\midrule
GEPA default & 27.88 & 28.08 & 28.27 & 28.85 &
55.81 & 56.29 & 56.14 & 57.33 \\
AdaGEPA & \textbf{31.92} & \textbf{34.04} & \textbf{31.92} & \textbf{32.31} &
\textbf{58.28} & \textbf{61.03} & \textbf{59.35} & \textbf{59.69} \\
\bottomrule
\end{tabular}
\end{table*}

At the matched $B=560$ checkpoint, AdaGEPA improved mean unseen-service
JGA by $5.96$ points, with a 95\% paired-seed interval of $[4.81,6.92]$.
At $B=1120$, the mean JGA differences were $+8.27$ points in-domain
(W/T/L $=5/0/0$, interval $[+4.23,+13.65]$) and $+3.46$ points on unseen
services (W/T/L $=4/0/1$, interval $[+0.58,+5.77]$). Comparing AdaGEPA
at $B=560$ with the GEPA default at $B=1120$ gives $+7.69$ points
in-domain and $+5.19$ points on unseen services, with W/T/L $=5/0/0$ for
both comparisons.

\subsubsection{SGD Extension Study}
\label{app:sgd-five-seed-extension}

The SGD extension study uses the same 5-formation configuration as
the initial study with a different set of paired optimization runs.
Table~\ref{tab:sgd-five-seed-extension} reports the paired differences
corresponding to the arm means in Table~\ref{tab:key-results-contrasts},
together with Gain-AUC and cross-budget comparisons. The cross-budget rows
compare AdaGEPA at $B=560$ with the GEPA default at $B=1120$.

\begin{table*}[t]
\centering
\small
\setlength{\tabcolsep}{4pt}
\renewcommand{\arraystretch}{1.07}
\caption{Paired SGD differences for the extension study.
All values are AdaGEPA minus the GEPA default, in points. The search-external
results use the existing fixed in-domain and unseen-service panels.}
\label{tab:sgd-five-seed-extension}
\begin{tabularx}{\textwidth}{@{}
>{\centering\arraybackslash}p{0.19\textwidth}
>{\centering\arraybackslash}X
ccc@{}}
\toprule
\multicolumn{1}{c}{\textbf{Evaluation}} &
\multicolumn{1}{c}{\textbf{Contrast and metric}} &
\multicolumn{1}{c}{\textbf{Mean}} &
\multicolumn{1}{c}{\textbf{W/T/L}} &
\multicolumn{1}{c}{\textbf{95\% paired-seed interval}} \\
\midrule
Development & Gain-AUC, Balanced DST & $+0.73$ & 3/0/2 & $[-1.06,+2.52]$ \\
Development & $B=560$ vs. $B=560$, Balanced DST & $+1.79$ & 4/0/1 & $[+0.03,+4.00]$ \\
Development & $B=1120$ vs. $B=1120$, Balanced DST & $-0.40$ & 3/0/2 & $[-1.67,+0.87]$ \\
\addlinespace
In-domain panel & $B=560$ vs. $B=560$, JGA & $+4.04$ & 3/0/2 & $[-2.12,+11.54]$ \\
In-domain panel & $B=1120$ vs. $B=1120$, JGA & $+1.54$ & 2/1/2 & $[-1.15,+4.23]$ \\
In-domain panel & $B=560$ vs. default $B=1120$, JGA & $-0.19$ & 2/0/3 & $[-3.65,+3.27]$ \\
\addlinespace
Unseen-service panel & $B=560$ vs. $B=560$, JGA & $+5.71$ & 5/0/0 & $[+3.72,+7.69]$ \\
Unseen-service panel & $B=1120$ vs. $B=1120$, JGA & $+0.77$ & 3/0/2 & $[-5.00,+6.54]$ \\
Unseen-service panel & $B=560$ vs. default $B=1120$, JGA & $+0.58$ & 3/0/2 & $[-3.91,+5.00]$ \\
\bottomrule
\end{tabularx}
\end{table*}

At $B=560$, the mean differences favor AdaGEPA on development and on both
search-external panels. The differences narrow by $B=1120$ as the default
search continues to improve.

\subsubsection{SGD Replication Study}
\label{app:sgd-ten-run-study}

The SGD replication study includes ten paired optimization runs under a fixed
10-formation configuration. Table~\ref{tab:sgd-ten-run-replication} reports
paired development and search-external differences at $B=560$ and $B=1120$,
together with comparisons between AdaGEPA at $B=560$ and the GEPA default at
$B=1120$. Development results use Balanced DST and Gain-AUC. Search-external
results use JGA or Balanced DST as indicated. All differences and intervals
are reported in points.

\begin{table*}[t]
\centering
\small
\setlength{\tabcolsep}{4pt}
\renewcommand{\arraystretch}{1.07}
\caption{Paired SGD differences for the replication study. Search-external
scores evaluate frozen checkpoint candidates on the fixed in-domain and
unseen-service panels.}
\label{tab:sgd-ten-run-replication}
\begin{tabularx}{\textwidth}{
    @{}
    >{\centering\arraybackslash}p{0.18\textwidth}
    >{\centering\arraybackslash}X
    c
    c
    c
    c
    @{}
}
\toprule
\multicolumn{1}{c}{\textbf{Evaluation}} &
\multicolumn{1}{c}{\textbf{Contrast and metric}} &
\multicolumn{1}{c}{\textbf{Mean}} &
\multicolumn{1}{c}{\textbf{Median}} &
\multicolumn{1}{c}{\textbf{W/T/L}} &
\multicolumn{1}{c}{\textbf{95\% paired-seed interval}} \\
\midrule
Development & Gain-AUC & $+0.99$ & $+0.11$ & 5/0/5 & $[-2.46,+4.72]$ \\
Development & $B=560$ vs. $B=560$, Balanced DST & $+1.58$ & $+0.51$ & 6/0/4 & $[-2.15,+5.86]$ \\
Development & $B=1120$ vs. $B=1120$, Balanced DST & $-0.03$ & $-0.24$ & 5/0/5 & $[-3.02,+2.90]$ \\
Development & $B=560$ vs. default $B=1120$, Balanced DST & $-1.61$ & $-1.48$ & 4/0/6 & $[-4.38,+1.12]$ \\
\addlinespace
In-domain panel & $B=560$ vs. $B=560$, JGA & $+5.58$ & $+5.29$ & 7/0/3 & $[-0.38,+11.44]$ \\
In-domain panel & $B=1120$ vs. $B=1120$, JGA & $+2.21$ & $+2.88$ & 7/0/3 & $[-1.92,+5.87]$ \\
In-domain panel & $B=560$ vs. default $B=1120$, JGA & $+1.25$ & $+4.81$ & 6/0/4 & $[-3.94,+6.06]$ \\
\addlinespace
Unseen-service panel & $B=560$ vs. $B=560$, JGA & $+2.02$ & $+1.92$ & 6/0/4 & $[-2.50,+6.73]$ \\
Unseen-service panel & $B=1120$ vs. $B=1120$, JGA & $-0.38$ & $+0.48$ & 5/0/5 & $[-3.65,+2.50]$ \\
Unseen-service panel & $B=560$ vs. default $B=1120$, JGA & $-1.35$ & $0.00$ & 5/0/5 & $[-5.48,+2.60]$ \\
Unseen-service panel & $B=560$ vs. $B=560$, Balanced DST & $+1.11$ & $+1.90$ & 6/0/4 & $[-2.86,+5.11]$ \\
Unseen-service panel & $B=1120$ vs. $B=1120$, Balanced DST & $-0.24$ & $+0.35$ & 5/0/5 & $[-3.23,+2.46]$ \\
Unseen-service panel & $B=560$ vs. default $B=1120$, Balanced DST & $-1.37$ & $-0.14$ & 5/0/5 & $[-4.98,+2.07]$ \\
\bottomrule
\end{tabularx}
\end{table*}

At $B=560$, the mean differences favor AdaGEPA on development and both
search-external panels. By $B=1120$, the development and unseen-service mean
differences are near zero, while the in-domain JGA difference remains positive
on average.

Table~\ref{tab:sgd-unseen-per-seed} reports per-run unseen-service differences
in Balanced DST and JGA for the initial, extension, and replication SGD
studies at $B=560$ and $B=1120$.

\begin{table*}[t]
\centering
\small
\setlength{\tabcolsep}{6pt}
\caption{Per-run SGD search-external differences on the unseen-service panel.
Values are AdaGEPA minus the GEPA default in percentage points.}
\label{tab:sgd-unseen-per-seed}
\begin{tabular}{@{}cccccc@{}}
\toprule
& & \multicolumn{2}{c}{\textbf{Balanced DST}}
& \multicolumn{2}{c}{\textbf{JGA}} \\
\cmidrule(lr){3-4}\cmidrule(l){5-6}
\textbf{SGD study} & \textbf{Run} & $\boldsymbol{B=560}$ & $\boldsymbol{B=1120}$
& $\boldsymbol{B=560}$ & $\boldsymbol{B=1120}$ \\
\midrule
Initial &
\begin{tabular}[c]{@{}c@{}}1\\2\\3\\4\\5\end{tabular} &
\begin{tabular}[c]{@{}c@{}}$+2.96$\\$+5.19$\\$+5.15$\\$+4.99$\\$+5.42$\end{tabular} &
\begin{tabular}[c]{@{}c@{}}$-0.38$\\$+3.46$\\$+1.05$\\$+5.46$\\$+2.23$\end{tabular} &
\begin{tabular}[c]{@{}c@{}}$+3.85$\\$+5.77$\\$+5.77$\\$+6.73$\\$+7.69$\end{tabular} &
\begin{tabular}[c]{@{}c@{}}$-1.92$\\$+5.77$\\$+2.88$\\$+6.73$\\$+3.85$\end{tabular} \\
\addlinespace
Extension &
\begin{tabular}[c]{@{}c@{}}1\\2\\3\\4\\5\end{tabular} &
\begin{tabular}[c]{@{}c@{}}$+1.07$\\$+5.10$\\$+3.92$\\$+6.89$\\$+1.37$\end{tabular} &
\begin{tabular}[c]{@{}c@{}}$+2.81$\\$-4.58$\\$+6.27$\\$-3.12$\\$+1.12$\end{tabular} &
\begin{tabular}[c]{@{}c@{}}$+3.53$\\$+7.05$\\$+5.77$\\$+9.29$\\$+2.88$\end{tabular} &
\begin{tabular}[c]{@{}c@{}}$+3.85$\\$-7.69$\\$+11.54$\\$-5.77$\\$+1.92$\end{tabular} \\
\addlinespace
Replication &
\begin{tabular}[c]{@{}c@{}}1\\2\\3\\4\\5\\6\\7\\8\\9\\10\end{tabular} &
\begin{tabular}[c]{@{}c@{}}$+5.10$\\$+0.14$\\$+7.74$\\$-6.01$\\$-5.42$\\$+3.66$\\$-8.13$\\$+11.02$\\$-4.53$\\$+7.53$\end{tabular} &
\begin{tabular}[c]{@{}c@{}}$+2.49$\\$-4.03$\\$+4.52$\\$-1.49$\\$-1.40$\\$+6.27$\\$-10.32$\\$+2.10$\\$+2.68$\\$-3.28$\end{tabular} &
\begin{tabular}[c]{@{}c@{}}$+8.65$\\$+0.96$\\$+9.62$\\$-6.73$\\$-5.77$\\$+2.88$\\$-6.73$\\$+13.46$\\$-5.77$\\$+9.62$\end{tabular} &
\begin{tabular}[c]{@{}c@{}}$+3.85$\\$-3.85$\\$+3.85$\\$-3.85$\\$-0.96$\\$+5.77$\\$-11.54$\\$+1.92$\\$+3.85$\\$-2.88$\end{tabular} \\
\bottomrule
\end{tabular}
\end{table*}

\subsubsection{SGD Formation-Prompt Comparisons}
\label{app:sgd-formation-prompt-comparisons}

\paragraph{Direct deployment.}
From five GEPA default formation searches, we selected the prompt with the
highest development score and evaluated it directly, without further
optimization. Table~\ref{tab:sgd-formation-direct-deployment} compares this
formation prompt with AdaGEPA's $B=560$ candidates on the same unseen-service
panel.

\begin{table}[t]
\centering
\small
\setlength{\tabcolsep}{5pt}
\caption{Direct deployment comparison on unseen-service SGD. Scores are JGA
percentages, and the difference is AdaGEPA minus the formation prompt in
percentage points.}
\label{tab:sgd-formation-direct-deployment}
\begin{tabular}{@{}ccccc@{}}
\toprule
\textbf{Evaluation} & \textbf{Formation prompt} &
\textbf{AdaGEPA} $\boldsymbol{B=560}$ & \textbf{Difference} & \textbf{95\% interval} \\
\midrule
Unseen-service JGA & 36.54 & 32.69 & $-3.85$ & $[-13.97,+5.82]$ \\
\bottomrule
\end{tabular}
\end{table}

The selected formation prompt scored above the mean of the AdaGEPA $B=560$
candidates, showing that formation search can itself yield a competitive
deployable prompt.

\paragraph{Continuation from the formation prompt.}
We also started the GEPA default and AdaGEPA from the same selected formation
prompt, allowing a paired comparison from a stronger initialization. By
$B=1120$, the GEPA default and AdaGEPA had improved on the starting prompt by
$3.22$ and $3.00$ Balanced DST points, respectively. AdaGEPA ended $0.22$
points below the default. Both methods therefore improved the stronger
starting prompt by similar amounts.

\subsubsection{MultiWOZ Search-External Checkpoint Evaluation}
\label{app:multiwoz-b560-external-supplement}

Frozen candidates saved at $B=560$ and $B=1120$ from five paired MultiWOZ
optimization runs were evaluated on a fixed 104-dialogue search-external panel
drawn from the official test split.
Table~\ref{tab:multiwoz-b560-external-per-seed} reports the per-run Balanced
DST and JGA differences at both budgets.

\begin{table*}[t]
\centering
\small
\setlength{\tabcolsep}{5pt}
\caption{MultiWOZ search-external differences on the fixed 104-dialogue panel
drawn from the official test split. Values are AdaGEPA minus the GEPA default
in percentage points.}
\label{tab:multiwoz-b560-external-per-seed}
\begin{tabular}{@{}ccccc@{}}
\toprule
& \multicolumn{2}{c}{\textbf{Balanced DST}}
& \multicolumn{2}{c}{\textbf{JGA}} \\
\cmidrule(lr){2-3}\cmidrule(l){4-5}
\textbf{Run} & $\boldsymbol{B=560}$ & $\boldsymbol{B=1120}$
& $\boldsymbol{B=560}$ & $\boldsymbol{B=1120}$ \\
\midrule
1 & $+3.30$ & $+3.49$ & $+2.88$ & $+3.85$ \\
2 & $-4.48$ & $+3.05$ & $-9.62$ & $-1.92$ \\
3 & $+6.11$ & $+3.67$ & $+8.65$ & $+4.81$ \\
4 & $+6.86$ & $+10.01$ & $+6.73$ & $+17.31$ \\
5 & $-4.01$ & $-4.64$ & $-8.65$ & $-9.62$ \\
\bottomrule
\end{tabular}
\end{table*}

At $B=560$, the mean and median Balanced DST differences were $+1.56$ and
$+3.30$ points, with W/T/L $=3/0/2$ and a 95\% interval of
$[-2.74,+5.85]$. The corresponding JGA differences were $0.00$ and $+2.88$
points, with W/T/L $=3/0/2$ and an interval of $[-6.73,+6.73]$. At
$B=1120$, the mean and median Balanced DST differences were $+3.12$ and
$+3.49$ points, with W/T/L $=4/0/1$ and an interval of $[-1.38,+7.32]$.
The corresponding JGA differences were $+2.88$ and $+3.85$ points, with W/T/L
$=3/0/2$ and an interval of $[-4.23,+10.96]$.

\subsubsection{HoVer Development and Search-External Evaluation}
\label{app:hover-search-external}

We evaluated stage-progress replay against the GEPA default across five paired
optimization runs. Table~\ref{tab:hover-per-seed} reports the development
endpoints and executed replay counts. Stage-progress replay improved the mean
endpoint by $2.47$ percentage points, with a median difference of $0.33$
points, W/T/L $=3/1/1$, and a 95\% interval of $[-0.60,+6.93]$ points. Replay
opportunities were sparse, but the candidates that passed the local gate
advanced the frontier and entered winner lineages.

\begin{table}[t]
\centering
\small
\setlength{\tabcolsep}{5pt}
\renewcommand{\arraystretch}{1.05}
\caption{Paired HoVer development results. Scores and differences are
percentages and percentage points, respectively. Replays count executed
stage-progress replacements.}
\label{tab:hover-per-seed}
\begin{tabular}{@{}ccccc@{}}
\toprule
\textbf{Run} & \textbf{GEPA default} & \textbf{AdaGEPA} & \textbf{Difference} & \textbf{Replays} \\
\midrule
1    & 43.00 & 43.00 & $0.00$   & 1 \\
2    & 43.33 & 43.67 & $+0.33$ & 2 \\
3    & 43.67 & 42.00 & $-1.67$ & 2 \\
4    & 38.33 & 49.67 & $+11.33$ & 4 \\
5    & 44.00 & 46.33 & $+2.33$ & 3 \\
\midrule
Mean & 42.47 & 44.93 & $+2.47$ & 2.4 \\
\bottomrule
\end{tabular}
\end{table}

Frozen candidates saved at $B=624$ and $B=1120$ from five paired HoVer
optimization runs were evaluated on a fixed 253-example search-external panel.
Table~\ref{tab:hover-search-external} reports the per-run differences in
Document Exact at both budgets.

\begin{table*}[t]
\centering
\small
\setlength{\tabcolsep}{4.5pt}
\renewcommand{\arraystretch}{1.06}
\caption{Per-run differences in HoVer search-external Document Exact. Values
are AdaGEPA minus the GEPA default on the fixed 253-example panel, in
percentage points.}
\label{tab:hover-search-external}
\begin{tabular}{@{}ccc@{}}
\toprule
\textbf{Run} & $\boldsymbol{B=624}$ & $\boldsymbol{B=1120}$ \\
\midrule
1 & $+1.98$ & $-0.79$ \\
2 & $-9.88$ & $+1.19$ \\
3 & $-1.98$ & $-6.32$ \\
4 & $+17.39$ & $+13.83$ \\
5 & $0.00$ & $+1.58$ \\
\bottomrule
\end{tabular}
\end{table*}

At $B=624$, the mean and median differences were $+1.50$ and $0.00$ points,
with W/T/L $=2/1/2$ and a 95\% interval of $[-5.53,+10.43]$. At $B=1120$,
they were $+1.90$ and $+1.19$ points, with W/T/L $=3/0/2$ and an interval of
$[-3.24,+8.46]$. On the secondary full 300-example panel at $B=1120$, the
mean difference was $+1.53$ points.

\subsection{What Drives Feedback Allocation?}
\label{app:allocation-controls}

The following controls examine four design choices: intervention capacity,
proposal support, task-specific targeting, and ordering. SGD provides the most
complete set of matched controls, while MultiWOZ, AIME, IFBench, and PUPA test
the corresponding choices under different feedback structures. HoVer and TACO
are included in the mechanism analysis in
Appendix~\ref{app:mechanism-funnel-counts}; the LiveBench-Math design study is
reported there as a feedback-pool saturation case.

\subsubsection{SGD: Full-Minibatch History and Coverage}
\label{app:sgd-full-minibatch-history-coverage}

The full-minibatch history-and-coverage sampler ranks feedback examples using
loss gaps, repeated failures, and coarse service coverage before resampling all
three minibatch positions.
Table~\ref{tab:sgd-history-coverage-matched-arm-means} compares it with the
GEPA default across five paired optimization runs.

\begin{table}[t]
\centering
\footnotesize
\setlength{\tabcolsep}{3.5pt}
\renewcommand{\arraystretch}{1.08}
\caption{Paired SGD development results for the full-minibatch
history-and-coverage sampler. Checkpoint values are Balanced DST percentages.
Gain-AUC is reported in points.}
\label{tab:sgd-history-coverage-matched-arm-means}
\begin{tabular}{@{}cccccc@{}}
\toprule
\textbf{Arm} &
$\boldsymbol{B=392}$ &
$\boldsymbol{B=560}$ &
$\boldsymbol{B=840}$ &
$\boldsymbol{B=1120}$ &
\textbf{Gain-AUC} \\
\midrule
GEPA default & 51.12 & 53.87 & 55.95 & 56.30 & 9.44 \\
History-and-coverage & 52.41 & 53.80 & 53.80 & 53.96 & 8.99 \\
\bottomrule
\end{tabular}
\end{table}

The sampler led the default by $1.29$ Balanced DST points at $B=392$. At
$B=560$, the two arms were nearly tied, and the default improved further at
later budgets. Across the five runs, 148 of 150 selected slots contained
first-time feedback occurrences, indicating that the rule mainly prioritized
examples not previously selected for reflection. The component metrics moved
in different directions. Active intent improved by $1.15$ points, while
requested-slot, slot-value, and joint-state exact scores decreased by $4.94$,
$1.02$, and $3.08$ points, respectively. This pattern motivated a narrower
intervention that preserves most of the scheduled minibatch and directs one
position using task-specific evidence.

\subsubsection{SGD: One-Slot Replacement and Ordering}
\label{app:sgd-development-controls}

The SGD controls examine whether a legal one-slot replacement is sufficient,
whether a task-structured proposal pool adds value, whether fitted ordering
adds value within that pool, and how results vary with the amount of formation
evidence. The table also includes neighboring feedback-allocation controls
based on failure history and representative scheduling.
Table~\ref{tab:additional-sgd-controls} reports paired differences for each
paired comparison. Each row uses its own paired set of optimization runs.
Results are not pooled across studies.

\begin{table*}[t]
\centering
\small
\setlength{\tabcolsep}{4pt}
\renewcommand{\arraystretch}{1.08}
\caption{Paired SGD development controls. Each entry is the first named arm
minus its comparator. Gain-AUC differences are reported in points. Endpoint
differences are reported in percentage points. Each row uses its own paired
run set.}
\label{tab:additional-sgd-controls}
\begin{tabularx}{\textwidth}{
    @{}
    >{\centering\arraybackslash}X
    >{\centering\arraybackslash}p{0.22\textwidth}
    >{\centering\arraybackslash}p{0.22\textwidth}
    @{}
}
\toprule
\multicolumn{1}{c}{\textbf{Contrast}} &
\multicolumn{1}{c}{\textbf{Gain-AUC / W/T/L}} &
\multicolumn{1}{c}{\textbf{Endpoint / W/T/L}} \\
\midrule
\multicolumn{3}{c}{\textbf{\textit{One-slot replacement and ordering}}} \\
AdaGEPA $-$ GEPA default &
$+3.97$ / $5/0/0$ &
$+3.92$ / $5/0/0$ \\
Structured Unordered $-$ GEPA default &
$+3.24$ / $3/0/2$ &
$+3.36$ / $3/0/2$ \\
AdaGEPA $-$ Structured Unordered &
$+0.74$ / $4/0/1$ &
$+0.56$ / $3/0/2$ \\
AdaGEPA $-$ Random Legal &
$+1.02$ / $5/0/5$ &
$+1.12$ / $5/0/5$ \\
Random Legal $-$ GEPA default &
$+0.03$ / $5/0/5$ &
$-0.83$ / $5/0/5$ \\
\addlinespace
\multicolumn{3}{c}{\textbf{\textit{Full-minibatch baseline and formation-size sensitivity}}} \\
AdaGEPA (5 formation runs) $-$ history-and-coverage &
$+0.27$ / $3/0/2$ &
$+1.90$ / $4/0/1$ \\
AdaGEPA (10 formation runs) $-$ history-and-coverage &
$+2.86$ / $3/0/2$ &
$+4.03$ / $3/0/2$ \\
AdaGEPA (10 formation runs) $-$ AdaGEPA (5 formation runs) &
$+2.59$ / $4/0/1$ &
$+2.13$ / $4/0/1$ \\
AdaGEPA (10 formation runs) $-$ GEPA default &
$+2.41$ / $3/0/2$ &
$+1.69$ / $3/0/2$ \\
AdaGEPA (15 formation runs) $-$ GEPA default &
$+1.60$ / $5/0/0$ &
$+0.48$ / $4/0/1$ \\
AdaGEPA (20 formation runs) $-$ GEPA default &
$-0.18$ / $3/0/2$ &
$-0.99$ / $2/0/3$ \\
\addlinespace
\multicolumn{3}{c}{\textbf{\textit{Neighboring feedback-allocation controls}}} \\
APEX-style failure prioritization $-$ GEPA default &
$+1.01$ / $3/0/2$ &
$+1.92$ / $3/0/2$ \\
AdaGEPA $-$ APEX-style failure prioritization &
$+2.97$ / $3/0/2$ &
$+2.00$ / $2/0/3$ \\
SESS-style representative schedule $-$ GEPA default &
$+2.17$ / $4/0/1$ &
$+2.73$ / $3/0/2$ \\
AdaGEPA (10 formation runs) $-$ SESS-style representative schedule &
$+0.24$ / $2/0/3$ &
$-1.05$ / $2/0/3$ \\
\bottomrule
\end{tabularx}
\end{table*}

Random Legal remained close to the GEPA default in these runs, so legal
one-slot replacement alone was not sufficient to reproduce AdaGEPA's mean
improvement. AdaGEPA had higher mean Gain-AUC and endpoint scores than Random
Legal. Structured Unordered retained most of AdaGEPA's mean improvement, while
adding fitted ordering produced a smaller mean increment in this configuration.
For AdaGEPA minus Random Legal, the 95\% paired-seed intervals were
$[-1.78,+3.63]$ Gain-AUC points and $[-1.79,+4.11]$ endpoint percentage
points. For AdaGEPA minus Structured Unordered, the corresponding intervals
were $[-1.89,+2.55]$ and $[-3.61,+4.85]$.

Beyond proposal-pool construction and ordering, we also isolate the
granularity of task-specific targeting. The service-level-only control targets
weak services but omits within-service refinement based on the parent's intent
and slot deficits.
Table~\ref{tab:sgd-service-only-control} compares its frozen candidates with
the GEPA default and AdaGEPA candidates from the initial SGD study on the
unseen-service panel. AdaGEPA exceeded the service-level-only control by
$4.81$ JGA points at $B=560$ and $0.38$ points at $B=1120$.

\begin{table}[t]
\centering
\small
\setlength{\tabcolsep}{9pt}
\caption{Unseen-service JGA (\%) for the initial SGD study. The
service-level-only control omits AdaGEPA's within-service component
refinement.}
\label{tab:sgd-service-only-control}
\begin{tabular}{@{}ccc@{}}
\toprule
\textbf{Arm} & $\boldsymbol{B=560}$ & $\boldsymbol{B=1120}$ \\
\midrule
GEPA default & 28.08 & 28.85 \\
Service-level-only & 29.23 & 31.92 \\
AdaGEPA & 34.04 & 32.31 \\
\bottomrule
\end{tabular}
\end{table}

APEX-style prioritizes feedback using failure history, while SESS-style places
a fixed representative subset at the beginning of each epoch. AdaGEPA had
higher mean Gain-AUC and endpoint scores than APEX-style in their paired
comparison. The 10-formation AdaGEPA configuration and SESS-style produced
similar mean Gain-AUC, while SESS-style had a slightly higher endpoint score.
Formation-size effects were nonmonotonic. The 10-formation configuration had
positive mean differences over the GEPA default in both Gain-AUC and endpoint
score. With 15 formation runs, the Gain-AUC difference remained positive but
the endpoint difference was small. With 20 formation runs, both mean
differences were negative.

\subsubsection{SGD: Same-State Allocation Comparisons}
\label{app:sgd-same-state-comparisons}

The comparison protocol and state selection are specified in
Appendix~\ref{app:sgd-same-state-protocol}. The four-arm panel evaluates the
GEPA default, Random Legal, Structured Unordered, and AdaGEPA with five
formation runs from the same saved SGD states. These comparisons measure
immediate frontier gain from the next proposal transition rather than outcomes
from a complete search trajectory.

Table~\ref{tab:sgd-reference-same-state} reports frontier-gain contrasts after
averaging three transition repeats within each state and then aggregating the
state-level values within each source optimization run. Random Legal had the
largest positive mean difference over the GEPA default. Structured Unordered
was close to Random Legal, while AdaGEPA did not increase mean frontier gain
over Structured Unordered on this panel.

\begin{table*}[t]
\centering
\small
\setlength{\tabcolsep}{4.5pt}
\renewcommand{\arraystretch}{1.08}
\caption{Four-arm same-state contrasts on saved SGD development states.
Frontier-gain differences and intervals are reported in percentage points.
Each source optimization run contributes three states, with three transition
repeats per distinct action and state.}
\label{tab:sgd-reference-same-state}
\begin{tabularx}{\textwidth}{
    @{}
    >{\centering\arraybackslash}X
    c
    c
    c
    c
    @{}
}
\toprule
\textbf{Contrast} &
\textbf{Mean} &
\textbf{Median} &
\textbf{W/T/L} &
\textbf{95\% paired-seed interval} \\
\midrule
AdaGEPA \(-\) Structured Unordered &
\(-0.187\) & \(-0.191\) & 2/0/3 & \([-0.454,+0.053]\) \\
Structured Unordered \(-\) Random Legal &
\(-0.007\) & \(-0.031\) & 2/0/3 & \([-0.641,+0.749]\) \\
Random Legal \(-\) GEPA default &
\(+0.312\) & \(+0.301\) & 4/0/1 & \([-0.022,+0.735]\) \\
AdaGEPA \(-\) GEPA default &
\(+0.118\) & \(+0.039\) & 3/0/2 & \([-0.573,+0.897]\) \\
\bottomrule
\end{tabularx}
\end{table*}

A second same-state panel compared AdaGEPA with 10 formation runs against its
corresponding Structured Unordered ablation, holding the proposal pool and
replacement opportunity fixed. Both arms executed 45 one-slot replacements.
The local gate passed for 41 AdaGEPA transitions and 42 Structured Unordered
transitions. Of these, 11 AdaGEPA transitions and 4 Structured Unordered
transitions advanced the current frontier.

\begin{table}[t]
\centering
\small
\setlength{\tabcolsep}{4pt}
\renewcommand{\arraystretch}{1.06}
\caption{Same-state frontier gains for the 10-formation AdaGEPA configuration.
Values are source-run means in percentage points after averaging three
transition repeats within each state.}
\label{tab:sgd-ten-run-same-state}
\begin{tabular}{@{}cccc@{}}
\toprule
\textbf{Source run} &
\textbf{AdaGEPA} &
\textbf{Structured Unordered} &
\textbf{Difference} \\
\midrule
1 & 0.000 & 0.144 & $-0.144$ \\
2 & 0.000 & 0.000 & 0.000 \\
3 & 1.739 & 0.167 & $+1.572$ \\
4 & 0.413 & 0.132 & $+0.281$ \\
5 & 0.958 & 0.000 & $+0.958$ \\
\midrule
Mean & 0.622 & 0.089 & $+0.533$ \\
\bottomrule
\end{tabular}
\end{table}

The paired mean difference was $+0.533$ percentage points, with a median of
$+0.281$, W/T/L $=3/1/1$, and a 95\% paired-seed interval of
$[-0.001,+1.135]$. With the replacement count and structured pool held fixed,
fitted ordering therefore produced a positive mean immediate gain on this
panel. The largest differences occurred in two of the five source runs.
Because the 5-formation and 10-formation panels use different saved states,
they show that the immediate contribution of ordering can vary across
configurations and state sets. They do not provide a matched comparison of
formation size.

\subsubsection{MultiWOZ: Service-Aware Routing and Ordering}
\label{app:multiwoz-allocation-controls}

Table~\ref{tab:multiwoz-development-controls} separates two choices in the
MultiWOZ adapter: whether replacement uses task-specific service-aware routing
rather than generic unused-example coverage, and whether eligible occurrences
within the selected service are ordered by a fitted model. Coverage-only
samples uniformly from unused feedback occurrences. The service-aware
rank-neutral arm retains service selection but removes fitted within-service
occurrence ordering. The fitted model was trained on MultiWOZ default-search
traces.

\begin{table}[t]
\centering
\small
\setlength{\tabcolsep}{5pt}
\renewcommand{\arraystretch}{1.05}
\caption{MultiWOZ component controls over five paired optimization runs.
Endpoint values are Balanced DST percentages, and Gain-AUC is reported in
points. Coverage-only removes task-specific service routing, while the
rank-neutral arm removes fitted occurrence ordering within the service-aware
proposal space.}
\label{tab:multiwoz-development-controls}
\begin{tabular}{@{}ccc@{}}
\toprule
\textbf{Arm} & \textbf{Endpoint} & \textbf{Gain-AUC} \\
\midrule
GEPA default                    & 53.59 & 11.46 \\
Coverage-only                   & 53.24 & 11.39 \\
Service-aware rank-neutral      & \textbf{55.39} & \textbf{12.94} \\
Service-aware + fitted ordering & 52.56 & 11.68 \\
\bottomrule
\end{tabular}
\end{table}

The service-aware rank-neutral arm exceeded the GEPA default by $1.79$
endpoint points and $1.48$ Gain-AUC points, with four of five paired runs
favoring it on both measures. Relative to Coverage-only, the corresponding
differences were $+2.15$ and $+1.55$ points, with paired intervals of
$[0.59,3.48]$ and $[0.08,3.09]$. Coverage-only remained close to the GEPA
default. Adding fitted within-service occurrence ordering reduced the endpoint
by $2.83$ points and Gain-AUC by $1.26$ points relative to the rank-neutral
arm. All five paired runs favored the rank-neutral arm on the endpoint. In
these runs, the improvement beyond coverage-only came from service-aware
routing rather than fitted within-service occurrence ordering.

\subsubsection{AIME: Full-Minibatch and One-Slot Controls}
\label{app:aime-capacity-controls}

Table~\ref{tab:aime-capacity-controls} examines whether the outcome of a
coarse history-and-coverage rule depends on how much of the feedback minibatch
it changes. The full-minibatch controls resample all three positions, whereas
the bounded variant changes at most one.

\begin{table}[t]
\centering
\small
\setlength{\tabcolsep}{7pt}
\renewcommand{\arraystretch}{1.06}
\caption{AIME development controls over five paired optimization runs.
The endpoint is the mean number of correctly solved validation problems out of
45 at $B=560$. Gain-AUC uses the same correct-answer scale.}
\label{tab:aime-capacity-controls}
\begin{tabular}{@{}ccc@{}}
\toprule
\textbf{Method} & \textbf{Gain-AUC} & $\boldsymbol{B=560}$ \textbf{endpoint} \\
\midrule
GEPA default                         & 7.33 & 20.4 \\
History-and-coverage, full minibatch & 5.45 & 18.8 \\
History-and-coverage, one slot       & 6.43 & 20.0 \\
APEX-style, full minibatch           & 5.54 & 18.6 \\
\bottomrule
\end{tabular}
\end{table}

Both full-minibatch controls trailed the GEPA default in endpoint quality and
Gain-AUC. For history-and-coverage, restricting the same selection rule to one
slot recovered $1.2$ correct answers and $0.98$ Gain-AUC units relative to
full-minibatch resampling. This reduced the endpoint gap from $1.6$ to $0.4$
answers, although the bounded variant remained below the GEPA default on both
summaries. The comparison therefore indicates that limiting intervention
capacity reduced the loss from coarse feedback selection, but did not make the
selection signal sufficient to improve the overall search.

\subsubsection{IFBench: Verifier Aggregation}
\label{app:ifbench-allocation-controls}

The IFBench development comparison examines two ways of aggregating
overlapping verifier feedback. Mean-deficit targeting averages the applicable
verifier deficits, whereas all-pass-risk targeting prioritizes examples by the
risk that at least one verifier fails. Table~\ref{tab:ifbench-three-arm}
compares their search outcomes.

\begin{table}[t]
\centering
\small
\setlength{\tabcolsep}{5pt}
\renewcommand{\arraystretch}{1.05}
\caption{IFBench development controls over five paired optimization runs.
Endpoints are strict all-verifier pass percentages at $B=560$, and differences
are relative to the GEPA default. Verifier-targeted replacements are pooled
across the five runs.}
\label{tab:ifbench-three-arm}
\begin{tabular}{@{}cccccc@{}}
\toprule
\textbf{Arm} &
\textbf{Endpoint} &
$\boldsymbol{\Delta}$ &
\textbf{Gain-AUC} &
$\boldsymbol{\Delta}$ &
\textbf{Targeted repl.} \\
\midrule
GEPA default  & 48.89 & --    & 2.92 & --    & -- \\
Mean-deficit  & 51.85 & +2.96 & 4.15 & +1.23 & 8  \\
All-pass-risk & 49.63 & +0.74 & 3.81 & +0.89 & 39 \\
\bottomrule
\end{tabular}
\end{table}

Mean-deficit targeting produced the larger endpoint and Gain-AUC improvements,
with paired W/T/L of $3/2/0$ and $4/1/0$, respectively. All-pass-risk targeting
executed more verifier-targeted replacements but produced a smaller endpoint
improvement. In this comparison, replacement frequency alone therefore does
not explain the search outcome. How the verifier signals were aggregated also
mattered.

\subsubsection{PUPA: Privacy Feedback Reuse}
\label{app:pupa-allocation-controls}

Observed-deficit replay reuses an eligible example for which privacy leakage
was previously observed. We compare it with the GEPA default and a
coverage-only one-slot control across five paired optimization runs
(Table~\ref{tab:pupa-three-arm-per-seed}).

\begin{table}[t]
\centering
\small
\setlength{\tabcolsep}{5pt}
\renewcommand{\arraystretch}{1.05}
\caption{PUPA development endpoints across five paired optimization runs.
Scores and differences are privacy--quality percentages and percentage
points. Coverage-only uses the same one-slot capacity but selects unseen
examples without using observed leakage feedback.}
\label{tab:pupa-three-arm-per-seed}
\begin{tabular}{@{}ccccc@{}}
\toprule
\textbf{Run} & \textbf{GEPA default} & \textbf{Coverage-only} & \textbf{Replay} &
\textbf{Replay} $\boldsymbol{-}$ \textbf{default} \\
\midrule
1   & 80.80 & 81.13 & 85.10 & $+4.30$ \\
2   & 86.94 & 86.75 & 84.59 & $-2.34$ \\
3   & 85.10 & 85.14 & 86.64 & $+1.54$ \\
4   & 83.64 & 86.40 & 90.75 & $+7.11$ \\
5   & 86.70 & 86.64 & 82.39 & $-4.31$ \\
\midrule
Mean & 84.64 & 85.21 & 85.90 & $+1.26$ \\
\bottomrule
\end{tabular}
\end{table}

Compared with coverage-only, observed-deficit replay increased the mean
endpoint by $0.69$ percentage points with W/T/L $=3/0/2$, and Gain-AUC by
$0.44$ points with W/T/L $=4/0/1$. The corresponding $t$ intervals were
$[-4.01,+5.39]$ and $[-2.01,+2.89]$ points. Compared with the GEPA default,
the mean endpoint increased by $1.26$ percentage points with W/T/L $=3/0/2$,
whereas Gain-AUC changed by $-0.11$ points with W/T/L $=2/0/3$. All 15
realized replay replacements used examples with previously observed privacy
leakage, while coverage-only used none. Compared with coverage-only, observed
leakage provided a positive mean replay signal, although its effect varied
across runs.

\paragraph{Prompt-component control.}
This control changes the functional components of the candidate prompt rather
than the feedback examples used for reflection. Its mean endpoint differed
from the GEPA default by $-0.13$ percentage points, with W/T/L $=2/0/3$.
Eligible component changes were supported in only two of five optimization
runs and executed in one, so we report this neighboring control separately
from the feedback-allocation comparisons.

\subsection{Where Allocation Value Persists or Disappears}
\label{app:mechanism-funnel-counts}

\subsubsection{Mechanism Funnel and Event-Level Evidence}

\begin{figure*}[t]
    \centering
    \includegraphics[width=0.9\textwidth]{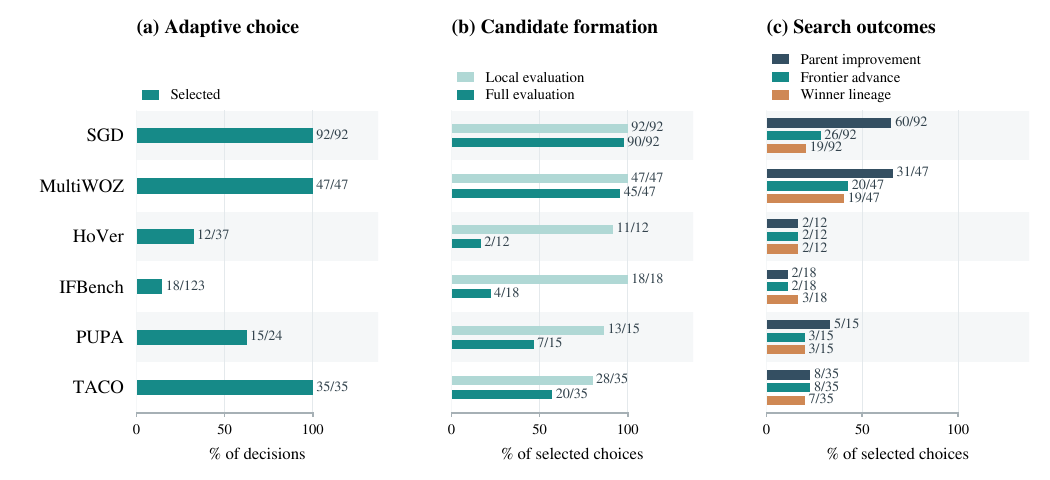}
    \caption{Feedback selection and candidate outcomes across six benchmarks.
    (a) Adaptive choices among feedback-allocation decisions.
    (b) Local and full evaluations following selected choices.
    (c) Parent improvements, frontier advances, and winner-lineage membership.
    Bars show percentages together with their counts and denominators.
    Panels (b) and (c) condition on the adaptive choices selected in panel (a).
    Counts are pooled across paired runs within each benchmark, and the outcomes
    in panel (c) may overlap.}
    \label{fig:mechanism-conversion}
\end{figure*}

The mechanism funnel traces an adaptive feedback choice from proposal to
candidate outcome. Nomination records a proposed replacement, support indicates
that it passed the eligibility checks, and action records an executed
replacement. After full evaluation, a parent improvement means that a candidate
outperforms its parent on the full validation set. A frontier advance raises
the best validation score found so far. Winner-lineage membership means that
the candidate lies on the parent--child path to the final winner, including
the winner itself. These outcomes may overlap and do not form a strictly
nested sequence. Figure~\ref{fig:mechanism-conversion} summarizes the
conversion rates, while Tables~\ref{tab:mechanism-funnel-formation}
and~\ref{tab:mechanism-funnel-outcomes} provide the corresponding counts.
Counts are pooled across paired runs for description, while statistical
comparisons retain the optimization run as their unit.

\begin{table*}[t]
\centering
\footnotesize
\setlength{\tabcolsep}{2.4pt}
\renewcommand{\arraystretch}{1.07}
\caption{Action realization and candidate formation in the reported search
comparisons. A dash means that the stage is not defined for that arm or that
its count cannot be reconstructed from the available logs.}
\label{tab:mechanism-funnel-formation}
\begin{tabular}{@{}cccccccccc@{}}
\toprule
\textbf{Task} & \textbf{Arm} &
\textbf{Dec.} & \textbf{Nom.} & \textbf{Sup.} & \textbf{Act} &
\textbf{Refl.} & \textbf{Local eval} & \textbf{Local pass} &
\textbf{Full eval} \\
\midrule
SGD & Service--component        & 92 & 92 & 92 & 92 & 92 & 92 & 90 & 90 \\
SGD & Random Legal               & 95 & 95 & 95 & 95 & 95 & 95 & 90 & 90 \\
SGD & default                    & 92 & -- & -- & -- & 92 & 92 & 90 & 90 \\
MultiWOZ & default               & 47 & -- & -- & -- & 46 & 46 & 45 & 45 \\
MultiWOZ & Service-aware routing & 47 & 47 & 47 & 47 & 47 & 47 & 45 & 45 \\
HoVer & default                  & 24 & -- & -- & -- & 23 & 23 & 10 & 10 \\
HoVer & Stage-progress replay   & 37 & -- & -- & 12 & 11 & 11 & 2 & 2 \\
IFBench & default                & 141 & -- & -- & -- & 135 & 135 & 28 & 28 \\
IFBench & Mean-deficit          & 123 & 18 & 18 & 18 & 18 & 18 & 4 & 4 \\
PUPA & default                   & 23 & -- & -- & -- & 19 & 17 & 12 & 12 \\
PUPA & Coverage-only             & 23 & 23 & 22 & 22 & -- & 17 & 13 & 13 \\
PUPA & Observed-deficit replay  & 24 & 24 & 15 & 15 & -- & 13 & 7  & 7  \\
TACO & default                   & 43 & -- & -- & -- & 36 & 36 & 20 & 20 \\
TACO & Coarse-skill             & 35 & 35 & 35 & 35 & 28 & 28 & 20 & 20 \\
\bottomrule
\end{tabular}
\end{table*}

\begin{table*}[t]
\centering
\footnotesize
\setlength{\tabcolsep}{4pt}
\renewcommand{\arraystretch}{1.07}
\caption{Outcomes of fully evaluated candidates in the same comparisons.
Parent improvements, frontier advances, and winner-lineage membership may
overlap. The final column counts candidates that both advanced the frontier
and entered the winner lineage.}
\label{tab:mechanism-funnel-outcomes}
\begin{tabular}{@{}ccccccc@{}}
\toprule
\textbf{Task} & \textbf{Arm} &
\textbf{Full eval} & \textbf{Parent gain} & \textbf{Frontier} &
\textbf{Lineage} & \textbf{Both} \\
\midrule
SGD & Service--component        & 90 & 60 & 26 & 19 & 17 \\
SGD & Random Legal               & 90 & 60 & 30 & 22 & 18 \\
SGD & default                    & 90 & 66 & 32 & 25 & 21 \\
MultiWOZ & default               & 45 & 31 & 19 & 16 & 12 \\
MultiWOZ & Service-aware routing & 45 & 31 & 20 & 19 & 17 \\
HoVer & default                  & 10 & 9 & 9 & 6 & 6 \\
HoVer & Stage-progress replay   & 2 & 2 & 2 & 2 & 2 \\
IFBench & default                & 28 & 10 & 6 & 8 & 4 \\
IFBench & Mean-deficit          & 4 & 2 & 2 & 3 & 2 \\
PUPA & default                   & 12 & 11 & 8  & 6  & 6  \\
PUPA & Coverage-only             & 13 & 9  & 7  & 5  & 4  \\
PUPA & Observed-deficit replay  & 7  & 5  & 3  & 3  & 3  \\
TACO & default                   & 20 & 9 & 6 & 5 & 4 \\
TACO & Coarse-skill             & 20 & 8 & 8 & 7 & 6 \\
\bottomrule
\end{tabular}
\end{table*}

Decision counts include steps at which no adaptive replacement was selected.
HoVer logs do not separate nomination from support.
Figure~\ref{fig:mechanism-conversion} uses decisions as the
denominator in panel (a) and selected adaptive choices in panels (b) and (c).

The funnels reveal different bottlenecks across tasks. SGD, MultiWOZ, and TACO
executed nearly all nominated replacements, so their differences arose after
the allocation decision was realized. MultiWOZ service-aware routing matched
the GEPA default in parent improvements and produced slightly more frontier
advances and winner-lineage candidates. HoVer and IFBench applied fewer
adaptive actions, while PUPA filtered choices both at support and during
candidate formation. In PUPA, seven replay-induced candidates passed the
local gate and three entered the final winner lineages.

Because the AIME controls resample all three minibatch positions rather than
making a one-slot replacement, we analyze their mechanism counts separately
from Figure~\ref{fig:mechanism-conversion}. History-and-coverage allocation and
APEX-style failure prioritization produced 23 and 26 parent improvements,
respectively, compared with 18 for the GEPA default, yet neither improved
endpoint quality or Gain-AUC. History-and-coverage advanced the frontier 13
times, but these advances summed to 36 validation points and occurred later in
the search. The GEPA default advanced the frontier 10 times for a total of 44
points. APEX-style produced 26 parent improvements compared with 18 for the
GEPA default, but the positive parent-relative gains summed to 75 points for
both arms. Event frequency alone therefore did not determine search quality.
The magnitude and timing of the gains explain the lower endpoint and Gain-AUC
of both controls.

\subsubsection{LiveBench-Math: Feedback-Pool Saturation}
\label{app:livebench-full-minibatch-precursor}

LiveBench-Math provides a small-feedback-pool setting for the full-minibatch
design study. The protocol uses 50 feedback examples, 50 search-validation
examples, and a rollout cap of $B=560$. The cluster-aware sampler combines loss
gap, repetition control, and coarse mathematical-topic coverage. Because it
can replace all three minibatch examples, this design differs from AdaGEPA's
one-slot intervention.

Table~\ref{tab:livebench-full-minibatch-curve} summarizes ten paired
optimization runs. Relative to the GEPA default, the cluster-aware sampler
improved the mean score by $0.4$, $0.8$, and $0.5$ correct answers at $B=150$,
$300$, and $560$, respectively, and increased mean Gain-AUC by $1.397$
correct-answer units. By $B=560$, the default schedule had covered $99.0\%$ of
the feedback pool on average. The diminishing lead suggests that feedback
allocation has less influence once the default schedule has exposed reflection
to nearly all available examples.

\begin{table}[!htbp]
\centering
\small
\setlength{\tabcolsep}{4pt}
\caption{LiveBench-Math development results over ten paired optimization runs.
Scores are the number of correct answers among 50 search-validation examples,
and Gain-AUC uses the same scale. Cluster-aware denotes the full-minibatch
design described above rather than the one-slot AdaGEPA method.}
\label{tab:livebench-full-minibatch-curve}
\begin{tabular}{@{}ccccc@{}}
\toprule
\textbf{Method} & $\boldsymbol{B=150}$ & $\boldsymbol{B=300}$ &
$\boldsymbol{B=560}$ & \textbf{Gain-AUC} \\
\midrule
GEPA default & 37.6 & 39.4 & 40.7 & 4.877 \\
Cluster-aware & 38.0 & 40.2 & 41.2 & 6.274 \\
\bottomrule
\end{tabular}
\end{table}

\clearpage
\subsection{Cross-Task Synthesis}
\label{app:task-outcome-summary}

Table~\ref{tab:task-level-outcome-summary} summarizes task-level outcomes
across evaluation settings, while
Table~\ref{tab:feedback-allocation-phase-map} relates each task's allocation
signal to the search behavior observed under that design.

\begin{table}[!htbp]
    \centering
    \scriptsize
    \setlength{\tabcolsep}{3.5pt}
    \renewcommand{\arraystretch}{1.00}
    \renewcommand{\tabularxcolumn}[1]{m{#1}}
    \caption{Task-level outcome summary. Differences are measured in each
    task's native metric relative to the GEPA default. The table distinguishes
    evaluation settings and intervention types. Results are not aggregated
    across tasks.}
    \label{tab:task-level-outcome-summary}
    \begin{tabularx}{\textwidth}{
        @{}
        >{\centering\arraybackslash}m{0.13\textwidth}
        >{\centering\arraybackslash}m{0.23\textwidth}
        >{\centering\arraybackslash}m{0.18\textwidth}
        >{\centering\arraybackslash}X
        @{}
    }
        \toprule
        \textbf{Task / intervention} &
        \textbf{Evaluation setting} &
        \textbf{Task metric} &
        \textbf{Summary relative to the GEPA default} \\
        \midrule
        SGD / one-slot &
        Three paired studies on frozen in-domain and unseen-service panels &
        JGA &
        At $B=1120$, in-domain differences were $+8.27$, $+1.54$, and
        $+2.21$ points across the initial, extension, and replication studies;
        unseen-service differences were $+3.46$, $+0.77$, and $-0.38$ points. \\
        \addlinespace
        MultiWOZ / one-slot &
        Development search and a frozen official-test panel &
        Balanced DST &
        Development endpoint $+1.79$ points; search-external $+3.12$ at
        $B=1120$. \\
        \addlinespace
        HoVer / one-slot &
        Development search and frozen search-external panel &
        Document Exact &
        Development endpoint $+2.47$ points and search-external endpoint
        $+1.90$ points at $B=1120$. \\
        \addlinespace
        AIME / full and one-slot &
        Five paired optimization runs &
        Correct answers out of 45 &
        The full-minibatch rule trailed the GEPA default by $1.6$ correct
        answers. Its bounded one-slot variant reduced this gap to $0.4$
        answers, while its Gain-AUC remained $0.89$ points lower. \\
        \addlinespace
        LiveBench-Math / full-minibatch &
        Ten paired optimization runs &
        Correct answers out of 50 &
        The cluster-aware sampler led by $0.5$ correct answers at $B=560$, compared
        with $0.8$ at $B=300$. \\
        \addlinespace
        IFBench / one-slot &
        Five paired optimization runs &
        Truncation-robust strict endpoint &
        Mean-deficit targeting $+2.96$ percentage points. \\
        \addlinespace
        PUPA / one-slot &
        Five paired optimization runs &
        Privacy--quality score &
        Observed-deficit replay $+1.26$ percentage points. \\
        \addlinespace
        TACO / one-slot &
        Five paired optimization runs &
        Mean-pass / strict pass &
        $+0.72/+1.09$ percentage points, with W/T/L $=3/0/2$ for both. \\
        \bottomrule
\end{tabularx}
\end{table}

\begin{table}[!htbp]
    \centering
    \scriptsize
    \setlength{\tabcolsep}{3.2pt}
    \renewcommand{\arraystretch}{1.00}
    \renewcommand{\tabularxcolumn}[1]{m{#1}}
    \caption{Cross-task interpretation of the signals used for feedback
    allocation and the search behavior observed in each task. These patterns
    motivate task-specific hypotheses rather than a single aggregate
    comparison across tasks.}
    \label{tab:feedback-allocation-phase-map}
    \begin{tabularx}{\textwidth}{
        @{}
        >{\centering\arraybackslash}m{0.112\textwidth}
        >{\centering\arraybackslash}m{0.28\textwidth}
        >{\centering\arraybackslash}X
        @{}
    }
        \toprule
        \textbf{Task} &
        \textbf{Allocation signal} &
        \textbf{Observed search behavior} \\
        \midrule
        SGD &
        Hierarchical service and component deficits &
        Structured replacement improves early discovery, while the advantage
        may narrow as the default schedule reaches more feedback. \\
        MultiWOZ &
        Service-aware routing &
        Selecting the relevant service helps, while fitted within-service
        ordering does not preserve the same gain. \\
        HoVer &
        Intermediate retrieval-stage progress &
        Replay opportunities are sparse, but some replay-induced candidates
        persist into winner lineages. \\
        AIME &
        Coarse clusters and failure history &
        Additional parent improvements do not consistently accumulate as
        frontier gains. \\
        LiveBench-Math &
        Coarse mathematical-topic clusters and loss gaps &
        The allocation advantage narrows as the default schedule approaches
        complete coverage of the small feedback pool. \\
        IFBench &
        Overlapping verifier deficits &
        A shared prompt update can redistribute gains across verifier
        objectives. \\
        PUPA &
        Observed privacy leakage &
        Leakage feedback provides an executable replay direction, although
        only some candidates remain useful through later search. \\
        TACO &
        Validation-supported primary-skill deficits &
        Bounded replacements are frequently executed, but only a subset
        advances the frontier or enters winner lineages. \\
        \bottomrule
    \end{tabularx}
\end{table}

\subsection{Resource Accounting}
\label{app:resource-accounting}

Table~\ref{tab:main-development-resources} reports per-run resource use for the
main one-slot development comparisons. The logical rollout budget $B$ counts
task-level evaluations, while requests, tokens, and configured cost record
physical usage. For comparable per-run totals, the shared initial evaluations
are assigned to both paired arms.

\begin{table}[!htbp]
\centering
\small
\setlength{\tabcolsep}{4.5pt}
\renewcommand{\arraystretch}{1.08}
\caption{Per-run resources at the reported development endpoints. Request,
token, and configured-cost entries are shown in AdaGEPA/GEPA-default order.
Tokens are reported in millions, and $n$ is the number of paired optimization
runs.}
\label{tab:main-development-resources}
\begin{tabular}{@{}cccccc@{}}
\toprule
\textbf{Task / study} & $\boldsymbol{B}$ & $\boldsymbol{n}$ &
\textbf{Successful requests} & \textbf{Tokens (M)} &
\textbf{Configured USD} \\
\midrule
SGD initial     & 1120 & 5  & 1107.2 / 1105.8 & 5.102 / 4.532 & 2.242 / 1.974 \\
SGD extension   & 1120 & 5  & 1110.4 / 1103.6 & 4.731 / 4.195 & 2.085 / 1.828 \\
SGD replication & 1120 & 10 & 1103.7 / 1104.4 & 4.728 / 4.590 & 2.059 / 2.005 \\
MultiWOZ       & 1120 & 5  & 1103.0 / 1104.4 & 4.723 / 5.072 & 2.050 / 2.209 \\
HoVer          & 1120 & 5  & 3775.8 / 3715.0 & 3.589 / 3.436 & 1.645 / 1.537 \\
IFBench        & 560  & 5  & 550.2 / 549.0   & 1.637 / 1.573 & 1.349 / 1.414 \\
PUPA           & 607  & 5  & 2781.8 / 2916.8 & 2.469 / 2.603 & 1.316 / 1.393 \\
TACO           & 1424 & 5  & 1323.4 / 1334.6 & 4.695 / 5.189 & 2.771 / 2.936 \\
\bottomrule
\end{tabular}
\end{table}

Table~\ref{tab:sgd-online-resource-comparison} compares the resources used to
reach the AdaGEPA candidate at $B=560$ with those used by the paired GEPA
default trajectory through $B=1120$ in each SGD study. The table reports
realized search-prefix usage, while candidate quality at the corresponding
checkpoints is reported separately in
Table~\ref{tab:key-results-contrasts}.

\begin{table}[!htbp]
\centering
\small
\setlength{\tabcolsep}{5pt}
\renewcommand{\arraystretch}{1.08}
\caption{Per-run physical resources for SGD search prefixes ending at AdaGEPA
$B=560$ and GEPA default $B=1120$. Entries are shown in that order, and tokens
are reported in millions. Usage pairs gives the number of runs with exact
prefix accounting relative to the total number of paired optimization runs.}
\label{tab:sgd-online-resource-comparison}
\begin{tabular}{@{}ccccc@{}}
\toprule
\textbf{SGD study} &
\textbf{Usage pairs} &
\textbf{Successful requests} &
\textbf{Tokens (M)} &
\textbf{Configured USD} \\
\midrule
Initial & 5/5 & 550.8 / 1105.8 & 1.869 / 4.532 & 0.826 / 1.974 \\
Extension & 5/5 & 551.4 / 1103.6 & 1.728 / 4.195 & 0.758 / 1.828 \\
Replication & 9/10 & 548.0 / 1104.6 & 1.807 / 4.627 & 0.783 / 2.022 \\
\bottomrule
\end{tabular}
\end{table}

Exact $B=560$ prefix accounting is available for nine of the ten replication
pairs, so the replication resource row uses those nine pairs. The quality
analysis includes all ten paired optimization runs.

Fitting the SGD ordering requires a formation set of prior GEPA-default
searches. The five-search formation set used 5,522 successful requests, 22.491
million recorded tokens, and \$9.816 in configured cost. The ten-search
formation set used 11,044 successful requests, 44.156 million tokens, and
\$19.168. These one-time formation resources are recorded separately from
subsequent optimization runs.

Table~\ref{tab:external-evaluation-resources} reports the physical resources
used for search-external evaluation of the frozen candidates in
Table~\ref{tab:key-results-contrasts}. These evaluation resources are recorded
separately from development search.

\begin{table}[!htbp]
\centering
\small
\setlength{\tabcolsep}{5pt}
\renewcommand{\arraystretch}{1.08}
\caption{Package-level resources for the reported search-external evaluations.
Requests count successful Task-LM calls, and tokens are reported in millions.}
\label{tab:external-evaluation-resources}
\begin{tabular}{@{}cccc@{}}
\toprule
\textbf{Task / study} & \textbf{Requests} & \textbf{Tokens (M)} &
\textbf{Configured USD} \\
\midrule
SGD initial & 4,264 & 19.579 & 7.136 \\
SGD extension & 5,824 & 21.461 & 7.902 \\
SGD replication, in-domain & 3,536 & 15.679 & 5.703 \\
SGD replication, unseen services & 3,536 & 15.607 & 5.692 \\
MultiWOZ, $B=560$ and $B=1120$ & 1,872 & 13.099 & 4.705 \\
HoVer, $B=624$ and $B=1120$ & 12,000 & 17.164 & 6.890 \\
\bottomrule
\end{tabular}
\end{table}

\end{document}